\documentclass[12pt]{article}

\usepackage[margin=1in]{geometry}
\usepackage{graphicx}
\usepackage{amsmath,amssymb}
\usepackage{authblk}
\usepackage[superscript]{cite}
\usepackage{hyperref}
\usepackage{siunitx}
\usepackage{xcolor}
\usepackage{stfloats}
\usepackage{placeins}
\usepackage{setspace}
\usepackage[scaled]{helvet}
\usepackage[font=small]{caption}
\usepackage{lineno}
\usepackage{booktabs}
\usepackage[T1]{fontenc}
\usepackage[utf8]{inputenc}
\fnbelowfloat

\makeatletter
\renewcommand{\maketitle}{%
  \begin{center}
    {\Large\bfseries \@title \par}
    \vskip 0.4em
    {\normalsize
      \lineskip .25em
      \begin{tabular}[t]{c}
        \@author
      \end{tabular}\par}
    \vskip 0em
  \end{center}
  \@thanks
}
\makeatother

\title{Evidence of an Emergent ``Self'' in Continual Robot Learning}
\author[1]{Adidev Jhunjhunwala\thanks{Correspondence to: aj3337@columbia.edu}}
\author[2]{Judah Goldfeder}
\author[1]{Hod Lipson}
\affil[1]{%
  Creative Machines Lab, Department of Mechanical Engineering, Columbia University, New York, NY
}
\affil[2]{%
  Creative Machines Lab, Department of Computer Science, Columbia University, New York, NY
}
\date{}

\begin{document}
\raggedbottom
\maketitle

% \modulolinenumbers[5]
% \linenumbers
\setlength{\affilsep}{0em}

\noindent\textbf{A key challenge to understanding self-awareness has been a principled way of quantifying whether an intelligent system has a concept of a ``self'', and if so how to differentiate the ``self'' from other cognitive structures. We propose that the ``self'' can be isolated by seeking the invariant portion of cognitive process that changes relatively little compared to more rapidly acquired cognitive skills - because our self is the most persistent aspect of our experiences. We used this principle to analyze the cognitive structure of robots under two conditions: One robot learns a constant task, while a second undergoes continual learning under variable tasks. We find that robots subjected to continual learning develop an invariant subnetwork that is significantly more stable ($p \ll 0.001$) compared to the control, and that this subnetwork is also functionally important: preserving it aids adaptation while damaging it impairs performance. We validate this pattern across three different robots spanning locomotion and manipulation.}
%We suggest that this principle can offer a window into exploring selfhood in other cognitive AI systems.

\enlargethispage{2\baselineskip}

\begin{figure}[!b]
  \centering
  \includegraphics[width=0.94\textwidth]{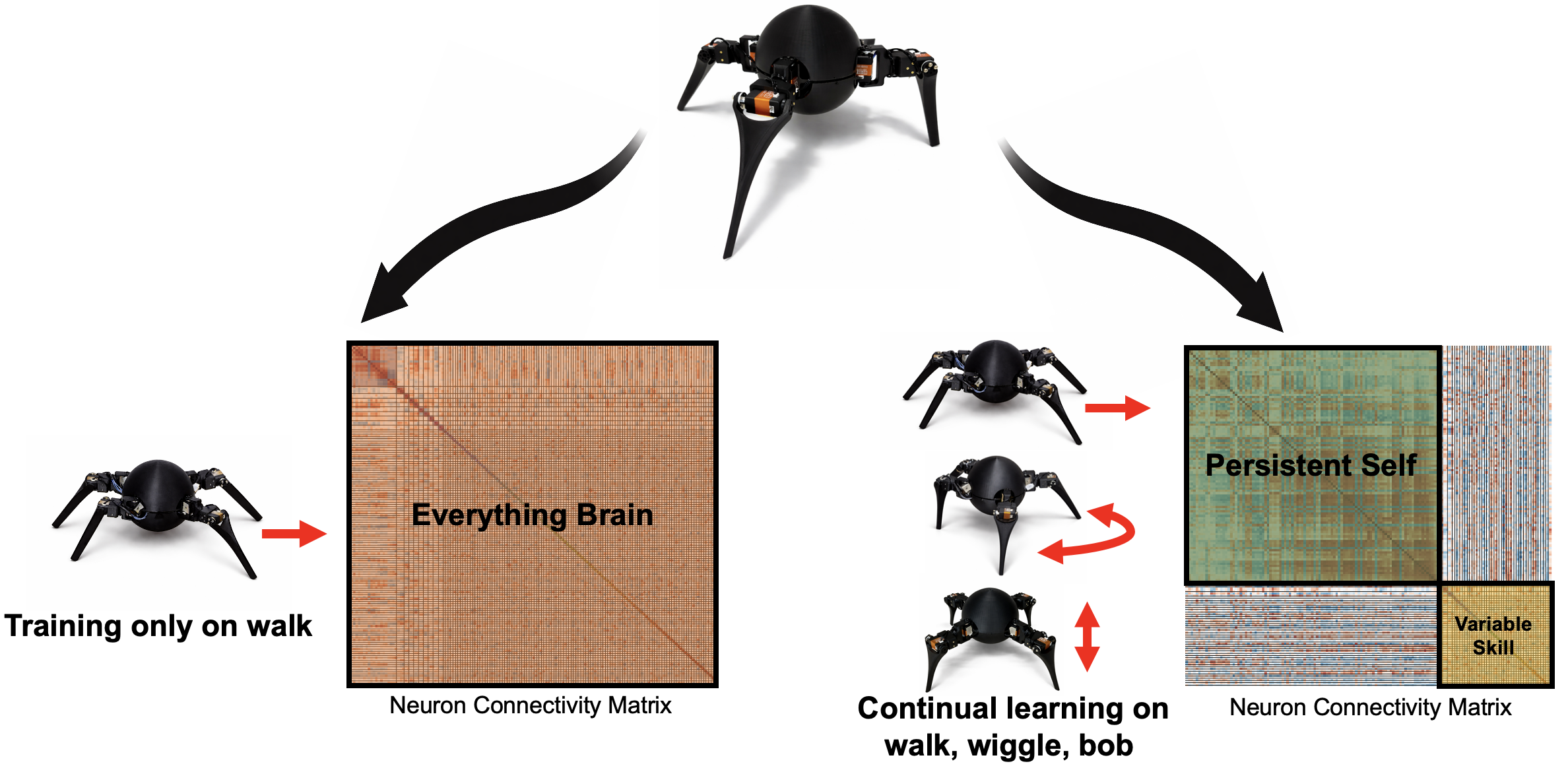}
  \caption{\textbf{Continual learning produces a stable self-like core.}
Compared to a single-task baseline, multi-behavior training yields a subnetwork that remains stable across behaviors (``persistent self''), while other components vary. Representative results from the first hidden layer of each network shown.}
  \label{fig:curriculum-learning}
\end{figure}

\clearpage
\raggedbottom
\setlength{\textfloatsep}{12pt plus 2pt minus 2pt}
\setlength{\footskip}{30pt}
% ---------- Main Text ----------
% \FloatBarrier

% \begin{figure}[!b]
%   \centering
%   \includegraphics[width=\textwidth]{Pictures/SimpleProc_Flat.png}
%   \caption{\textbf{Continual learning produces a stable self-like core.}
% Compared to a single-task baseline, multi-behavior training yields a compact subnetwork that remains identifiable and stable across behaviors (``persistent self''), while other components vary. Only the first hidden layer of each network is shown for clarity.}
%   \label{fig:curriculum-learning}
% \end{figure}
% ---------- INTRODUCTION ----------
\section{Introduction}
\label{sec:introduction}
 
A central question for philosophy, and more recently robotics and AI, is: What is the inner signature of selfhood? We hypothesize that one way to isolate the ``self'' is to look for the portion of the mind that is invariant across lifelong experiences. Consider two human bodies thrown into a swimming pool: one has spent decades walking, running, and interacting with the world; the other has only ever done one task. Both must now learn to swim, but only one, by virtue of its varied experiences, has learned to separate task-specific knowledge from knowledge that is invariant across tasks. This gives it a rich, implicit model of its own body~--- its kinematic limits, useful muscle synergies, and typical responses to motor commands, and thus we expect it to learn to swim much faster. The only commonality to every task is the agent itself; we therefore call this invariant knowledge the "self". We seek to test this idea in robots: whether, in a continual learning setting, a learned controller represents only task-specific action rules, or whether it also contains a more persistent subnetwork that remains stable across tasks because it encodes the body itself rather than the task at hand.

The self has been defined in many ways, and no single formula is universally accepted. However, many important accounts converge on one core idea: selfhood involves continuity through change. Locke linked personal identity to continuity across time, describing a person as ``the same thinking thing in different times and places''~\cite{locke1975essay}. Ricoeur similarly argued that identity is not merely strict sameness, but the preservation of selfhood through change~\cite{ricoeur1992oneself}. Parfit argued that what matters may not be a hidden self at all, but the continuity that is preserved across changing mental states~\cite{parfit1984reasons}. Merleau-Ponty, by contrast, stressed that even a basic sense of self is grounded in the lived body and in action in the world~\cite{merleauponty2012phenomenology}. For robots, this suggests that the best candidate for a minimal self is the part of the controller that remains relatively stable across changing behaviors while still helping organize action in the same body over time.

This definition is intentionally modest. We do not claim evidence of human-like consciousness. Rather, we ask whether, when the same robot body learns different tasks, part of its controller behaves less like a task-specific solution and more like a stable, reusable structure tied to the agent itself. In this sense, we define a minimal embodied self not by introspection, but by continuity through change in the same physically grounded agent.
 
Despite impressive performance, deep reinforcement learning for robot locomotion policies are often treated as monolithic functions whose internal representations are difficult to disentangle. There is also a growing body of work on self-modeling and body-schema learning in robots; however, most of that work explicitly segregates the robot's self-model from the rest of the controller, using a variety of techniques. For example, some self-modeling approaches explicitly learn an internal model of the robot's morphology that is continuously updated to support resilience to damage and change~\cite{bongard2006resilient}. More recent methods reconstruct the robot's body visually~\cite{chen2022full}, discover morphology from physical interaction data~\cite{diazledezma2023self}, or infer a body schema from exteroceptive and proprioceptive signals~\cite{jiang2024robot,pugach2019brain}. Our hypothesis is that under certain conditions, training across multiple behaviors could induce a persistent shared internal structure~\cite{lipson2002origin_variation,clune2013evolutionary_modularity} to promote efficient learning.
 
There is already evidence that multi-task learning (simultaneously training different tasks) motivates shared representations for generalization and transfer~\cite{caruana1997multitask}, and RL work has explored when and why sharing is beneficial~\cite{borsa2016sharedmtrl,deramo2020sharing}. But when behaviors are trained sequentially, we enter a continual-learning regime, where distribution shift induces interference and can lead to catastrophic forgetting, determining what is retained versus overwritten~\cite{khetarpal2022crlreview,kirkpatrick2017ewc,rusu2016progressive}. This tension makes multi-behavior locomotion an appealing testbed: if a stable, behavior-invariant representation of the body exists, it should be the part that survives behavior switches, while behavior-specific components reorganize.
 
To investigate this hypothesis, we probed a robot's ``self'' inside a standard deep learning policy by training a single controller across multiple distinct behaviors and comparing which internal components remain stable and which reorganize to learn a new behavior. We trained simulated robots across three morphologies to perform distinct behaviors in a cyclical sequence --- a condition we refer to as \emph{continual learning}. For the two locomoting bodies (a quadruped and a hexapod), these behaviors were walk, wiggle, and bob; for a robotic manipulation arm~\cite{wolczyk2021continualworld,yu2020metaworld}, they were a sequence of distinct manipulation task families. We then identified subnetworks by grouping co-activated neurons and measuring how consistently those groups persisted across cycles, comparing the continual-learning policies to single-task controls trained for an equal number of cycles.
 
Our results reveal strong evidence of a persistent group of neurons (corresponding to a subnetwork) that remains identifiable across changing tasks. This subnetwork is more stable than the corresponding groupings observed under the constant task scenario. Alongside this persistent group, there are other more plastic groups of neurons that reorganize and change to adapt to the variable-task condition. This interpretation is also supported by intervention: when the self subnetwork is frozen during further learning, performance is better preserved than when an equally sized task-like subnetwork is frozen ($p<0.01$, 594 matched pairs), whereas lesioning the self causes a larger performance drop than lesioning the task subnetwork. We further replicate this pattern in a robotic manipulation setting, showing the finding is not limited to locomotion or to the specific morphologies.

\section{Results}
\label{sec:results}
\begin{figure}[t]
  \centering \includegraphics[width=\textwidth]{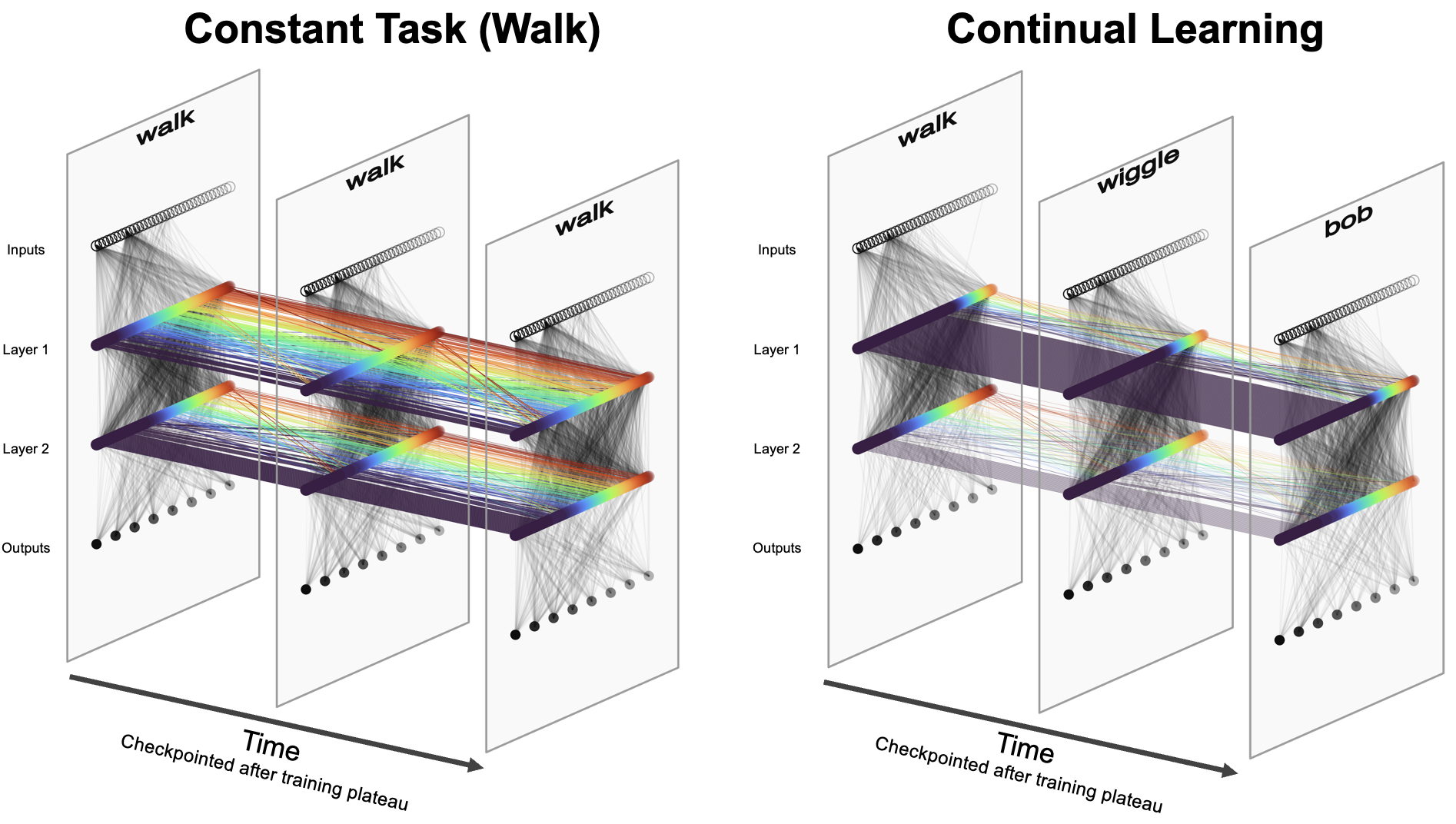}
  \caption{\textbf{Visualization of the persistent self in static and variable conditions.} Each policy is shown on its own plane, with hidden-layer units ordered by co-activation-based subnetworks, showing how the size and structure of the subnetworks change across learning. Alluvial flows connect matched neuron families across successive policies, grouped by source subnetwork $\rightarrow$ target subnetwork; dark purple denotes the self subnetwork. Flow width indicates how many matched families pass between subnetworks, and opacity encodes mean persistence score. \textbf{Left:} three consecutively trained walk policies produce a more fragmented organization, with flows distributed across many groupings. \textbf{Right:} walk$\rightarrow$wiggle$\rightarrow$bob reveals one dominant subnetwork that remains continuous across policies, while smaller groupings split, merge, and reroute more strongly. Results shown for the quadruped.}
  \label{fig:alluvial-sketch}
\end{figure}

\subsection{Learning behaviors}

Before analyzing internal representations, we confirmed that continual learning reliably produces three distinct, repeatable behaviors in a single quadruped policy. Within each training cycle, each phase reaches a stable solution and at each switch we transfer only the policy weights to initialize the next phase, yielding consistent walk, wiggle (in-place rotation), and bob (vertical hopping) behaviors under identical morphology, training parameters, and dynamics. We also validated these behaviors on a physical quadruped to ensure that the learned behaviors are kinematically valid in physical reality (Sec.~\ref{app:readyant}). We reproduced the same walk--wiggle--bob curriculum on a hexapod morphology, confirming that both the behaviors and the downstream analysis are not specific to the quadruped's body plan (Sec.~\ref{app:hexapod}). To test whether the finding extends beyond locomotion, we also replicated the full pipeline on a robotic manipulation benchmark across four task families (Sec.~\ref{app:continual-world-results}).

For the experimental control, we trained constant-task policies — walk-only, wiggle-only, or bob-only — for the same total number of training phases as the continual-learning policies, and used the resulting checkpoints as baselines. We repeated this entire experiment in eight independent runs with different random seeds to serve as a constant task baseline.

\subsection{A persistent ``self''-like subnetwork}
\label{sec:results-self}
 
We tested our hypothesis on the plateaued walk, wiggle, and bob networks of the quadruped and compared the resulting structure to a single-task baseline. Figure~\ref{fig:alluvial-sketch} provides an intuitive view of the structure we tested for: whether the neural network contains a stable, reusable, task-agnostic grouping of neurons that remains relatively unchanged as the behavior changes, versus a fully entangled representation in which task-related and body-related structure are not separable. To test this hypothesis, we first grouped connected neurons by creating a co-activation matrix that captures when pairs of neurons' activations are correlated~\cite{horta2021coactivation,weil2025coactivationtransformers}. We then used a standard block diagonalization procedure~\cite{CuthillMcKee1969Bandwidth,GeorgeLiu1981SparsePD} to find the major blocks of neurons in this matrix, corresponding to the subnetwork. Finally, we checked whether these blocks (subnetworks) remain persistent across time, or they reorganize. Our hypothesis is that under changing conditions (continual learning), there will be a persistent subnetwork (the ``self'') clearly separable from the rest of the neurons (which are more task-specific), whereas under static conditions there will be just one large group of neurons (the ``everything brain'', Fig~\ref{fig:curriculum-learning}).

\begin{figure}[p]
  \centering
  \begin{minipage}{0.95\textwidth}
    \centering
    \includegraphics[width=0.95\textwidth]{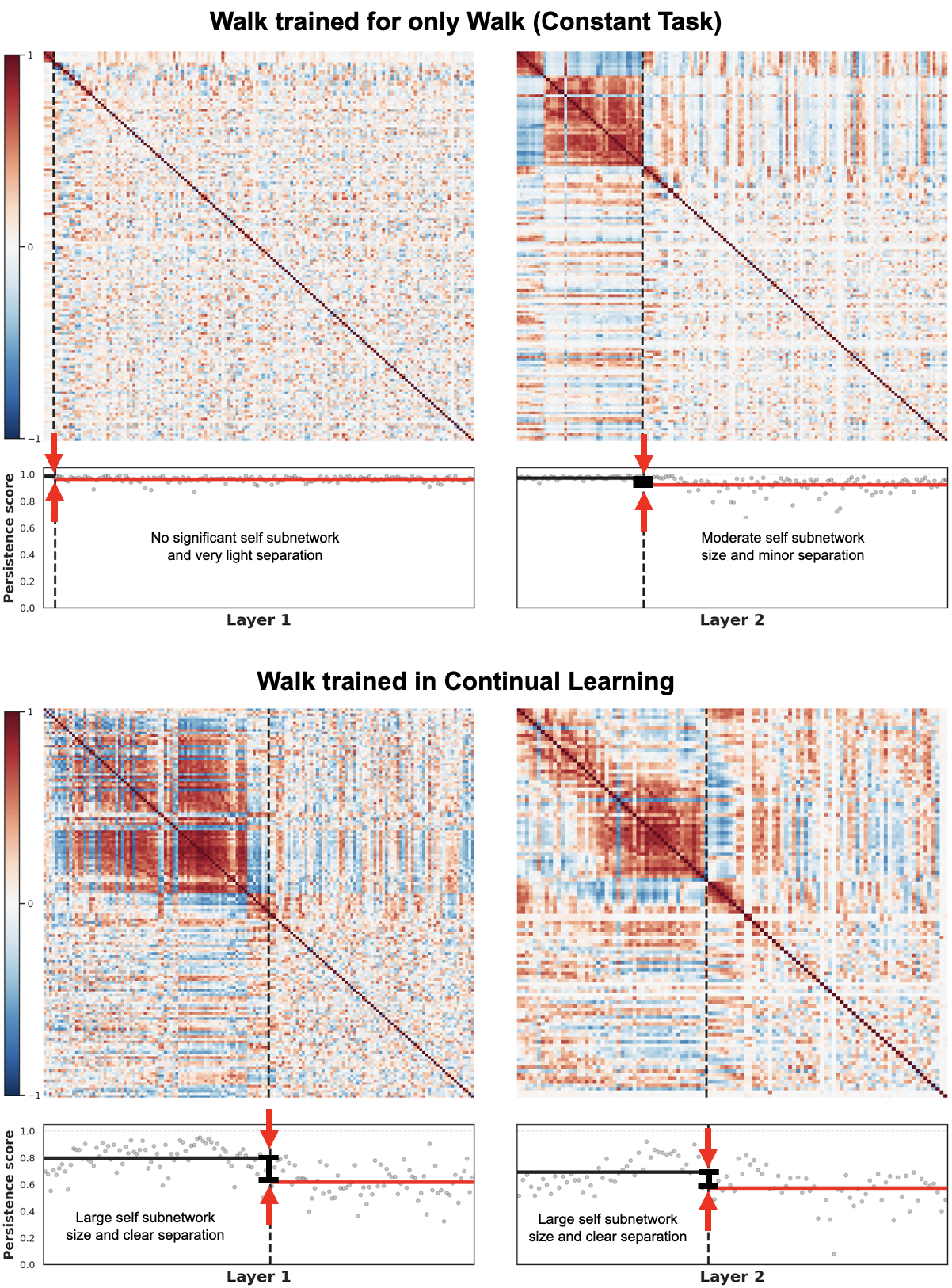}\\[6pt]
  \end{minipage}
  \caption{\textbf{Quantitative evidence for a persistent self-like subnetwork.} Shown here is one trained policy from each condition (constant-task and continual). Although both policies successfully perform the same behavior, they exhibit markedly different internal structure. The top panel shows the reordered neuron--neuron co-activation matrix with inferred subnetwork boundaries, and the bottom panel shows per-neuron persistence score in the same ordering. For a run-level overlay of this view across many plateaued snapshots and additional examples, see Sec.~\ref{app:tessellation-and-examples}. Results shown for the quadruped.}
  \label{fig:analysis-quant}
\end{figure}

For a walk-only baseline, cross-policy connections are visually diffuse. When we compare consecutively trained walk checkpoints, the networks do not yield a clean, trackable self subnetwork, and matches fragment into many small, scattered streams. Consistent with this qualitative picture, the quantitative diagnostics in Figure~\ref{fig:analysis-quant} show no clear separation into a dominant high-stability subnetwork; instead, persistence scores are distributed more smoothly across subnetworks, suggesting that once performance has saturated, additional training primarily produces weak, noisy reorganization rather than refining a stable core.

By contrast, in the continual-learning policy, both the sketch and the quantitative view reveal a markedly different picture. In Figure~\ref{fig:alluvial-sketch} (right), a dominant first-layer subnetwork remains stable across walk, wiggle, and bob: neurons in the largest subnetwork stay grouped, and the cross-policy flows concentrate into a strong, stable band rather than dispersing broadly across the whole network. Figure~\ref{fig:analysis-quant} makes the same separation explicit: the largest subnetwork in each hidden layer exhibits substantially higher mean persistence score than the smaller groupings, indicating a compact subnetwork whose co-activation and connectivity structure are preserved across behaviors. This is the signature of a self-like subnetwork in our sense: \emph{a subset of the controller that persists as the robot switches what it is doing, consistent with a reusable representation of the body --- a minimal cognitive self that persists beneath changing skills,} while other parts reorganize more strongly to implement behavior-specific control.

A key question is whether this persistent subnetwork is merely a stable chunk of the network, or whether it plays an important and generalizable role. We tested this by performing two intervention analyses during transfer to a new behavior. Across matched transfer trials evaluated from the end of the warmup period until plateau, freezing the self subnetwork led to better performance than freezing an equally sized task subnetwork. Conversely, lesioning the self subnetwork led to worse performance than lesioning the task subnetwork. These effects were consistent across the paired comparisons and in the expected direction in both analyses, with strong resampling-based support in the direct matched comparisons (freeze: 92.7\%; lesion: 91.3\%). When the same comparisons were evaluated using a within-transition rank-normalized analysis, confidence exceeded $99\%$; full intervention details are provided in Supplementary Materials~\ref{app:freeze-lesion}. Together, these intervention results show that the persistent subnetwork is not only more stable across behavior switches, but also functionally more important for cross-behavior adaptation.
 
\subsection{Self emergence and persistence across layers and cycles}
\label{sec:results-cycles}
 
\begin{figure}[t]
  \centering
  \includegraphics[width=\linewidth]{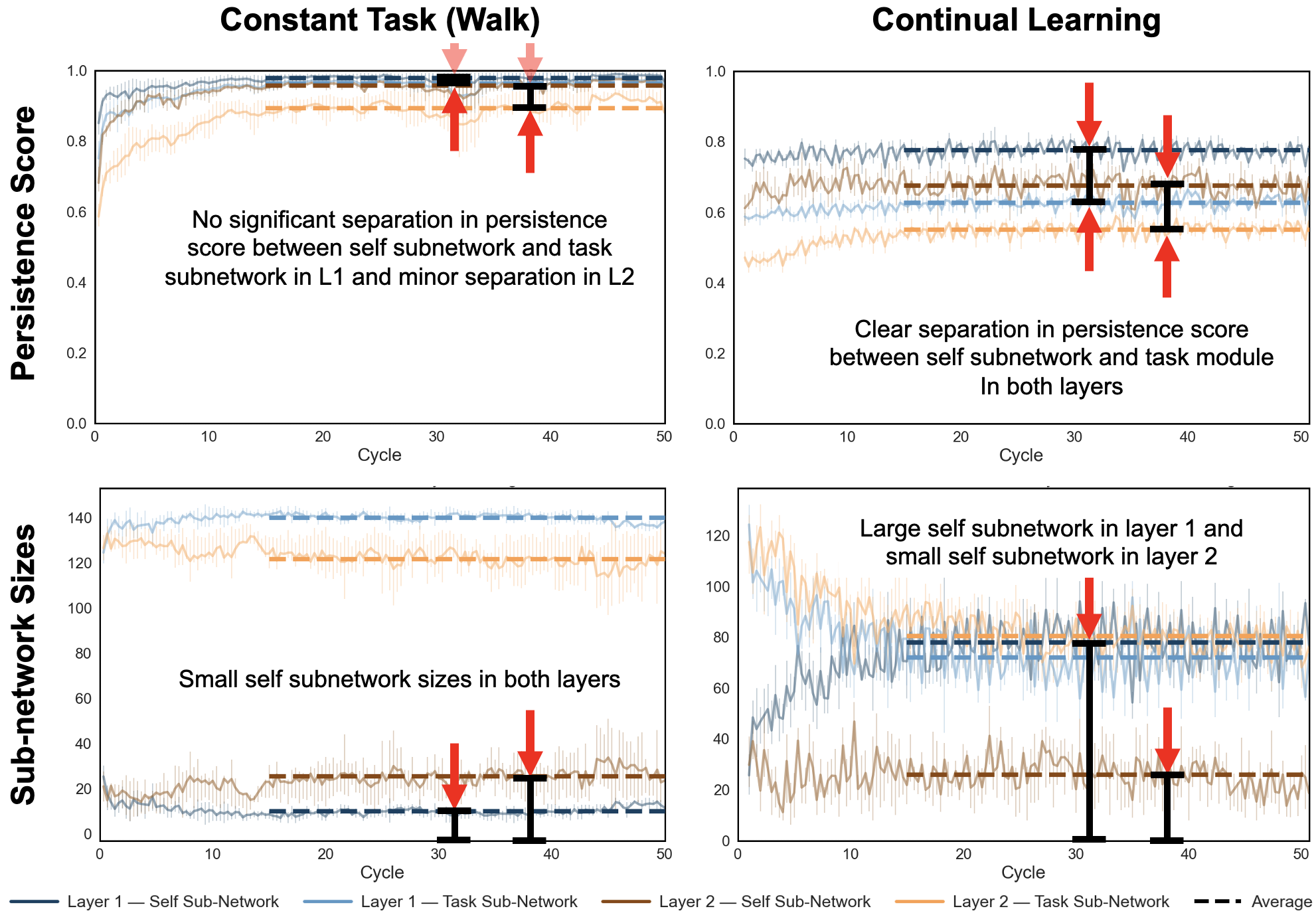}
  \caption{\textbf{Subnetwork persistence and size: constant-task baseline vs.\ continual-learning.} Mean \emph{persistence score} (top) and self subnetwork size (bottom) across 50 cycles for both hidden layers. The continual-learning (multi-behavior) agent shows a clear separation between the self subnetwork (largest subnetwork) and the pooled task subnetwork (all remaining units), while the walk-only baseline exhibits weaker separation and comparatively small self-subnetwork sizes. Error bars indicate inter-quartile range (IQR); dashed lines denote across-cycle means. Results shown for the quadruped.}
  \label{fig:cluster-evolution}
\end{figure}
 
Having identified and functionally validated a self-like subnetwork, we tested whether this structure is sustained across training, how it emerges over training cycles, where in the network it is most clearly expressed, and whether it reflects continual multi-behavior learning rather than a generic byproduct of optimization. Figure~\ref{fig:cluster-evolution} tracks both subnetwork persistence and size across training cycles while comparing continual-learning against a walk-only control.
 
In the continual-learning policy (averaged across 10 independent seeds), the self subnetwork (defined as the largest co-activation subnetwork in a layer) consistently exhibits a higher mean persistence score than the pooled task subnetwork across layers and cycles, i.e., it changes less when learning new behaviors. The persistence gap is therefore not limited to the specific walk--wiggle--bob example shown in Figures~\ref{fig:alluvial-sketch} and~\ref{fig:analysis-quant}; rather, it is sustained with a mean separation of 16.9 percentage points at 99\% confidence (Sec.~\ref{app:pstats}) across all seeds. Figure~\ref{fig:cluster-evolution} also shows how this structure develops over training: under continual learning, the number of neurons in the self subnetwork grows early and then plateaus, suggesting that as the policy evolves repeatedly, it grows its internal representation of itself. In the quadruped's $150{\times}150$ architecture, the self subnetwork saturates at roughly $\sim$80 neurons in layer~1 and $\sim$30 neurons in layer~2, while the pooled task subnetwork plateaus at roughly $\sim$60 neurons in layer~1 and $\sim$80 neurons in layer~2. The smaller effective self-subnetwork size in the later layer is partly attributable to dead ReLU units (near-zero activation variance on the shared reference states), which are excluded from our analysis. Across all continual-learning trials, this qualitative pattern is consistent: earlier layers tend to develop a much larger self-like subnetwork, while later layers retain a smaller but still identifiable stable core alongside more behavior-dependent structure. The stable subnetwork is not completely static---its activity still adapts to the current behavior---but it changes noticeably less than the rest of the network, preserving a core of ``who the robot is'' while allowing other components to flexibly implement ``what the robot is doing.''
 
The walk-only baseline (averaged across eight independent seeds) provides an important control. The baseline exhibits much weaker and less consistent separation between the largest subnetwork and the pooled remainder, especially in layer~1 where the persistence scores nearly overlap, with only a modest separation in layer~2. Just as importantly, the inferred self subnetwork sizes in the walk-only condition remain very small in both layers, never expanding into a substantial shared core. Thus, the walk-only setting does not show the combination of strong persistence separation and substantial self subnetwork recruitment seen under continual learning, indicating that the self--task separation arises specifically under continual multi-behavior switching, where some portion of the network preserves reusable body-related structure while other components reorganize to support the current behavior.
 
These conclusions are further supported by run-level overlays across plateaued checkpoints, included in Supplementary Materials~\ref{app:tessellation-and-examples}, which show that the same self-like subnetwork remains visible throughout the curriculum rather than appearing only at isolated transitions. In addition, we tested a set of alternative actor architectures and training variants---including widths from 100 to 250 units per layer, 1-, 2-, and 3-layer models, ELU, ReLU, and tanh activations, and normalization enabled versus disabled---and observed the same qualitative separation: a dominant self subnetwork with comparatively high persistence score, alongside more task-sensitive structure that is less stable across behaviors. The same separation was observed in the hexapod and in the manipulation replication across all four task families (Supplementary Materials~\ref{app:continual-world-results}), validating that this finding is not specific to the quadruped or to locomotion.

\section{Discussion}
\label{sec:discussion}

\begin{figure}[t]
  \centering
  \includegraphics[width=\linewidth]{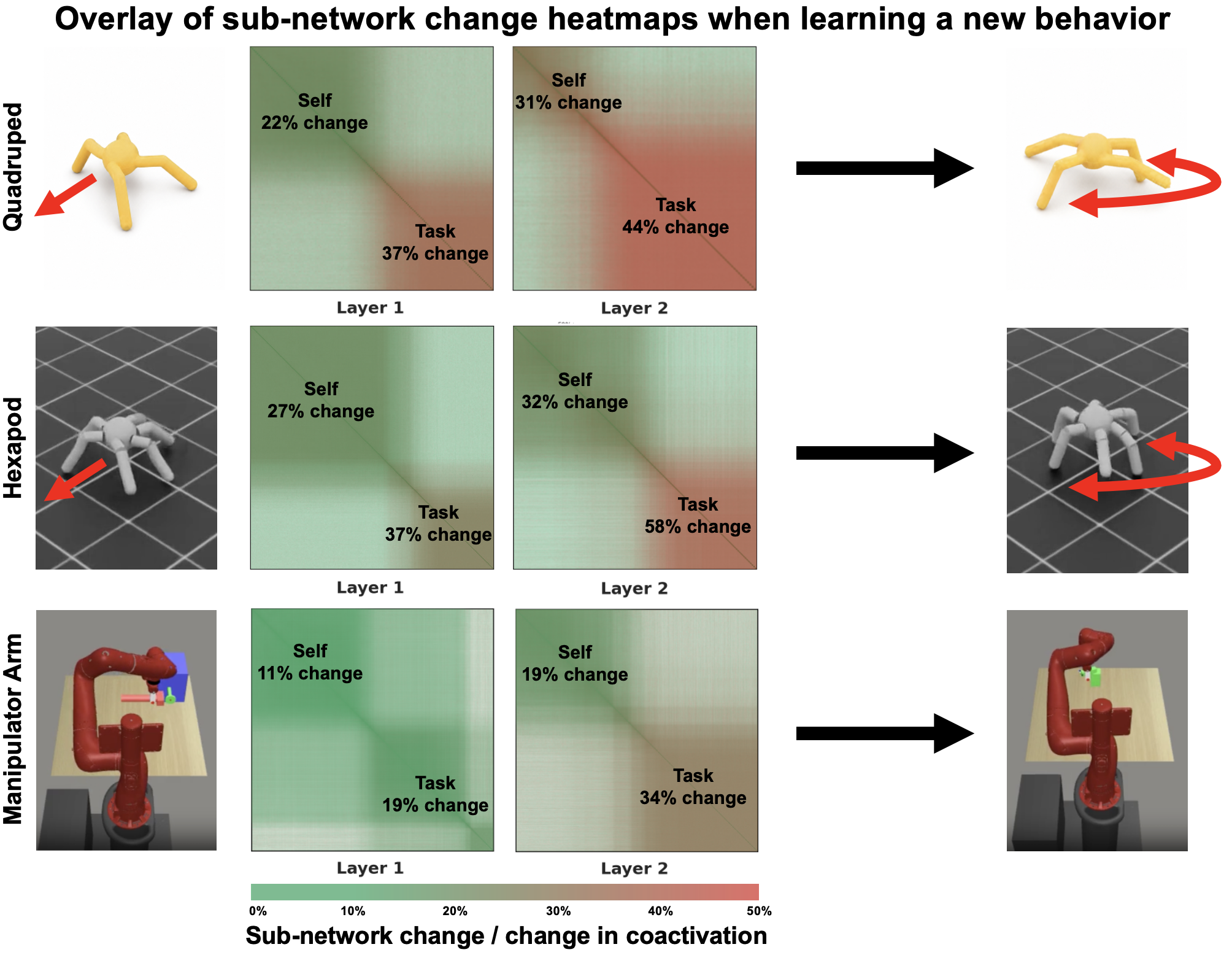}
  \caption{\textbf{Reorganization concentrates in task-like regions at behavior switches.}
  Overlay heatmaps show how much each subnetwork changes when learning a new behavior. Rows show the three robot systems; for each, the first two hidden layers are displayed. The self-like subnetwork exhibits consistently smaller change than the pooled task-like region in every system. The arm's network is deeper, so only its first two layers are shown for comparability. (See Sec ~\ref{app:continual-world-results} for all four layers)}
  \label{fig:delta-overlay}
\end{figure}

What could it mean for a robot to have a self? There is no true introspection to appeal to and no inner report to take at face value. The only handle left is the body and how it behaves over time: a self, in that setting, is what remains stable in the same body as the behavior itself changes. This is the definition we adopt, and our results meet it. The persistent subnetwork we identify is the part of the controller that remains stable as the agent walks, wiggles, and bobs --- satisfying Locke's and Ricoeur's criterion that identity is what continues through change --- while still helping the same body act across all three behaviors, consistent with Merleau-Ponty’s emphasis on bodily organization for action. A generic shared backbone would satisfy the first of these but not the second; a task-specific circuit would satisfy neither. What we observe satisfies both, and it does so without any self-modeling component or architectural constraint~\cite{bongard2006resilient,chen2022full,diazledezma2023self}. We do not claim that this amounts to self-reflective awareness or consciousness. We claim only that it satisfies the minimal definition of a self that is available to a robot: continuity through change in the same physical agent.

Figure~\ref{fig:delta-overlay} makes this reorganization visually explicit: across layers, weight changes concentrate outside the dominant self subnetwork, consistent with a stable core supported by a more plastic task substrate. That this core is not merely slow-changing but also functionally important suggests it reflects something structurally persistent about the agent's embodiment, rather than being an artifact of optimization pressure alone.

The separation between self-like and task-like subnetworks suggests practical ways to work with learned policies. Representation stability offers a concrete diagnostic for which parts of a controller encode reusable body dynamics versus behavior-specific control, instead of treating the policy as a monolith~\cite{puiutta2020xrlsurvey,milani2022xrlsurvey,acero2024interpretablelocomotion,lomasov2025chess_alignment}. A self-like subnetwork identified this way could serve as a backbone when learning new behaviors, or be monitored during fine-tuning to detect when body-related representations are being overwritten~\cite{rusu2016progressive,fernando2017pathnet,kirkpatrick2017ewc}. More generally, the spontaneous emergence of a modular, reusable core is consistent with broader principles of scalable system design~\cite{lipson2007principles_modularity}, and raises the question of whether explicitly exploiting this structure could improve transfer efficiency or robustness to physical perturbations such as joint damage or morphological change.

The main pattern reported here is robust across a range of seeds, actor architectures, training variants, and morphologies: earlier layers consistently exhibit the strongest cross-behavior invariance, while later structure is more behavior-linked~\cite{yosinski2014transferable}. The behavior set used here is intentionally compact: walking, wiggling, and bobbing cover much of the qualitatively distinct whole-body repertoire available to this morphology --- forward translation, in-place reorientation, and vertical impulse generation --- analogous to distinct temporally extended skills~\cite{sutton1999options}. The same persistence pattern replicates in a robotic manipulation setting across multiple task families, suggesting the finding is not specific to any one morphology or control domain. A natural next step is to test whether the self-like subnetwork remains stable and becomes explicitly useful as behavior suites expand and agents are deployed on hardware, particularly for agents whose bodies may be repaired, extended, or reconfigured over time~\cite{wyder2025robot_metabolism}.

Methodologically, we use co-activation based grouping and cross-behavior alignment as a simple, transparent way to expose modular organization in dense policies. Co-activation structure will not capture every aspect of functional role, but it provides a reproducible handle on groups of neurons that work together across shared reference states. In our setting, this is important because neurons that repeatedly co-activate across behaviors are not encoding a single behavior-specific output, but are likely helping coordinate a stable, body-centered representation of the agent's own state and dynamics, while other units reorganize to implement the current skill~\cite{horta2021coactivation,weil2025coactivationtransformers}. Future work may add complementary tools such as causal interventions, sparsity- or factorization-based decompositions, and targeted ablations to more directly test the necessity and sufficiency of specific subnetworks.

In summary, when the same agent is pushed to express different skills sequentially, part of its controller behaves less like a task solution and more like a stable description of the agent itself --- persistent across behaviors and functionally important during adaptation. This provides a concrete, testable notion of ``self'' in deep control: not an added module or explicit model, but a structure that emerges from the pressure to remain the same agent across changing demands, and that may reflect a general principle of how cognitive systems organize selfhood.

\section{Methods}
\label{sec:methods}

\begin{figure}[t]
  \centering
  \includegraphics[width=0.9\textwidth]{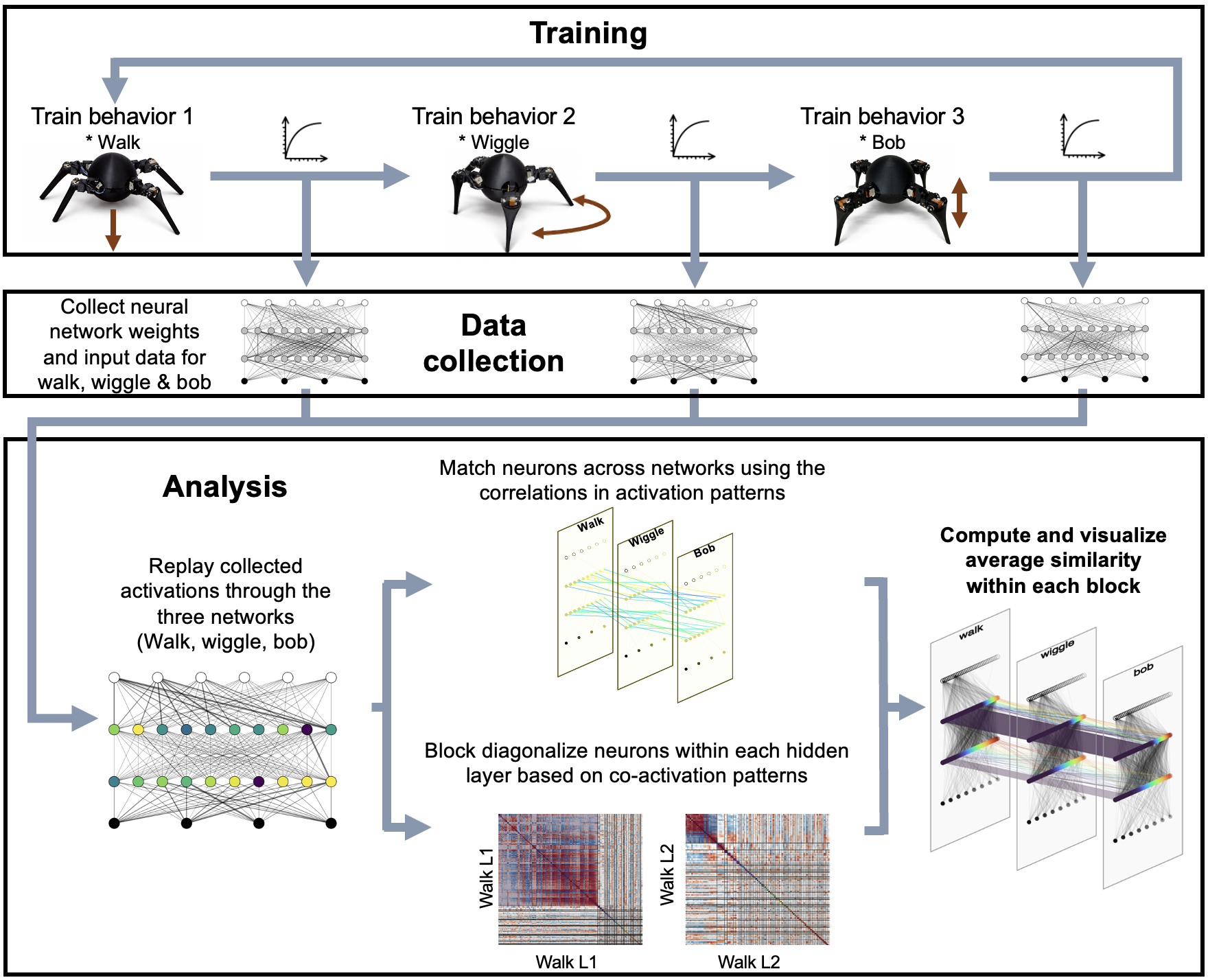}
  \caption{\textbf{Overview of the training and comparison pipeline in the multi-behavior experiment.} A single agent morphology was trained sequentially on walk, wiggle, and bob using SAC, with actor weights transferred between phases and plateau-based phase switching. After training, we compared the resulting policies on a shared set of reference states to identify neural groups that remained stable across behaviors and neural groups that reorganized more strongly. Shown for the quadruped; the same pipeline was applied to the hexapod and, with task families in place of locomotion behaviors, to the manipulation arm.}
  \label{fig:methods-overview}
  \vspace{6pt}
\end{figure}

\subsection{Training setup and curriculum}

For each morphology, we trained a single agent to perform behaviors sequentially under a fixed curriculum. Within this curriculum, each behavior was treated as a separate training phase with its own replay buffer and step budget. At each transition, we carried over only the policy weights. The learning algorithm and hyperparameters were identical in every phase; the only change was the reward function (Sec.~\ref{app:behaviors}).

For the primary experiments, we used the standard IsaacLab Quadruped (Ant-like) robot---four legs with two actuated joints per leg (eight actuated degrees of freedom)~\cite{gymnasium_ant,mittal2023isaaclab}---with IsaacLab's default configuration on a single flat ground plane with fixed dynamics, without modifying link masses, joint limits, friction, or gravity, so the self-like structure we identified could not be attributed to finely tuned environment choices. We also replicated the full pipeline on a hexapod (Sec.~\ref{app:hexapod}) and on a manipulation benchmark, training a robotic arm sequentially across multiple Meta-World task families using a separate architecture and training configuration~\cite{wolczyk2021continualworld,yu2020metaworld}; details are provided in Supplementary Materials~\ref{app:continual-world-results}.

We trained all robots with Soft Actor--Critic (SAC)~\cite{haarnoja2018sac,haarnoja2018sacapps}. For the quadruped, both actor and critic were 2-layer MLPs with 150 hidden units per layer and ReLU activations; the hexapod and manipulation arm used separate architectures and training configurations (Sec.~\ref{app:rl-details}). Training was performed in parallel vectorized simulation with fixed-horizon episodes and timeout-only termination.

Phase switching was controlled by plateau detection on episodic returns. To avoid switching too early, we enforced a minimum training period before enabling plateau checks. Once enabled, we compared returns in two back-to-back sliding windows and declared a phase converged when performance was both (i) stable, in the sense that the mean return changed only marginally between windows, and (ii) above a behavior-specific minimum threshold that we validated by visually inspecting rollouts. To prevent unproductive training, if a phase failed to reach the minimum threshold within a step budget, we treated it as failed, reverted to the previous checkpoint, and retried the phase with a new random seed (up to two retries). Full plateau-detection parameters and an example rollout are provided in Supplementary Materials~\ref{app:plateau}.

\subsection{Analysis of the persistent and task-specific neural structure}

We asked which hidden units in the policy behaved like a reusable ``sense of self'' across behaviors, and which were more task-specific. To answer this, we took actor networks trained on consecutive behavior phases and passed them through a shared set of reference states spanning all behaviors. Using shared reference inputs is consistent with the general practice of representation comparison across networks via activation statistics~\cite{raghu2017svcca,kornblith2019cka}.

Figure~\ref{fig:analysis-quant} shows the core visual diagnostics used in this comparison: the reordered co-activation matrix with inferred subnetwork boundaries~\cite{eisen1998cluster,horta2021coactivation}, and per-neuron persistence score in the same ordering. We used these outputs to identify self-like subnetworks with high stability and to track how those subnetworks evolved across cycles and training runs as shown in Figure~\ref{fig:cluster-evolution}.

\textbf{Observations and hidden activations.} For each trained behavior policy, we generated rollouts and collected observation vectors, then pooled observations by taking one full successful episode from each model and sampling a fixed-size $T$ shared reference set from this pooled buffer. All subsequent comparisons evaluated every policy on this identical input set to remove differences caused by behavior-specific state visitation. We then forwarded the shared observations through the actor network and recorded hidden-unit activations at the first two fully connected layers. For each layer, we normalized each unit's activation trace using a per-unit z-score; units with near-zero variance were treated as dead units and excluded from downstream analyses.

\textbf{Co-activation matrices and block diagonalization.} For each behavior and hidden layer, we computed a neuron--neuron cosine similarity matrix over the shared reference states: treating each neuron as a vector of activations across $T$ states, we set $R_{ij}=\cos(a_i,a_j)$, producing an $H\times H$ signed co-activation matrix $R$. As in prior work that treated similarity structure over unit activations as an interpretability primitive~\cite{horta2021coactivation,weil2025coactivationtransformers}, we used the absolute value $\lvert R\rvert$ as a nonnegative affinity matrix. We identified subnetworks via \emph{block diagonalization}: thresholding $\lvert R\rvert$ at $\tau$ to build a sparse ``strong-similarity'' graph and defining subnetworks as the resulting connected neuron groups. For visualization, we reordered neurons using \emph{reverse Cuthill--McKee (RCM)} ordering~\cite{CuthillMcKee1969Bandwidth,GeorgeLiu1981SparsePD} to make blocks visually compact. We fixed $\tau$ at 70\% for all main results; results were invariant to threshold choice across a broad range (Sec.~\ref{app:k_sensitivity}).

\textbf{Neuron families (cross-behavior matching).} To track ``the same'' neuron across behaviors, we aligned units by solving an optimal one-to-one assignment problem, necessary because hidden-unit identities are only defined up to permutation symmetries~\cite{entezari2022permutation,ainsworth2023gitrebasin,simsek2021symmetries}. We applied the Hungarian algorithm~\cite{kuhn1955hungarian,munkres1957assignment} to cosine-similarity matrices between unit activation vectors across neighboring behavior networks, yielding a permutation that defined neuron families consistently aligned across behaviors.

\textbf{Persistence score.} For each neuron family $k$, we quantified two types of cross-behavior consistency and defined the persistence score as their average: an activation-similarity term $\text{act\_sim}_k$ (cosine similarity between that family's activation vectors across behavior pairs) and a connectivity-similarity term $\text{conn\_sim}_k$ (cosine similarity between the $k$-th row of the family--family co-activation matrix across behavior pairs).

\textbf{Subnetwork-level aggregation (self-like vs.\ task-like).} We defined subnetworks as co-activation based groups within each hidden layer and computed mean persistence score by averaging over all neuron families within each subnetwork. The largest subnetworks consistently exhibited the highest mean persistence scores. To summarize checkpoint-to-checkpoint reorganization, we converted persistence to \emph{percent change} ($100\%\times(1-\text{persistence})$) and averaged within each subnetwork, yielding an intuitive contrast between stable self-like and plastic task-like blocks during behavioral acquisition.

\subsection{Freezing and lesioning the self and task subnetworks}
\label{sec:methods-interventions}

To test whether the persistent subnetwork is functionally important for cross-behavior adaptation, we performed two intervention analyses during transfer to a new behavior: \emph{freezing} and \emph{lesioning}. In both analyses, we selected a fixed number of neurons and applied the same-sized intervention either to the self subnetwork or to the task subnetwork. In the freezing analysis, we froze all incoming weights to the selected neurons during further learning~\cite{kim2022continual,rusu2016progressive}. In the lesion analysis, we reinitialized the selected neurons while preserving their overall statistics, disrupting their learned function without changing the size of the affected region~\cite{bau2020units,morcos2018importance}. Comparisons were made on matched transfer runs sharing the same transition type, random seed, run ID, and source training cycle, so that the intervention target was the primary difference between paired runs. We report both direct paired reward differences and a within-transition rank-normalized comparison~\cite{wilcoxon1945}, which reduces the effect of reward-scale differences across transitions.

\bibliographystyle{naturemag}
\bibliography{refs}

\clearpage

% \section*{Implementation details}
% Each seed was trained on one RTX 2080 Ti; typical phase (1–2h) at 8192 envs

\section*{Code Availability}
All code is available at \url{https://github.com/adidevj7/EmergentRobotSelf}. The repository includes the Isaac-based training pipeline, the full analysis toolkit used in this study, configuration files, and checkpoints. The Isaac runtime is not redistributed due to licensing; the repository README provides setup instructions for obtaining the required NVIDIA components.

\section*{Data Availability}
Processed data (checkpoints, training logs, analysis outputs, statistics, and intervention results) supporting this study are available at \url{https://github.com/adidevj7/EmergentRobotSelf}.

\section*{Funding}
This work was supported in part by the US National Science Foundation (NSF) AI Institute for Dynamical Systems (DynamicsAI.org) under grant 2112085.

\section*{Author Contributions}
H.L. and A.J. conceived the project. A.J. led experiment design and implementation, developed the training and analysis pipelines, and conducted the experiments. J.G. and H.L. provided technical direction and feedback on experimental design, analysis methodology, and interpretation. A.J. drafted the manuscript, and A.J., J.G., and H.L. revised and edited the manuscript.

\section*{Competing Interests}
The authors declare no competing interests.

\clearpage
\begin{center}
  {\Large\bfseries Supplementary Materials for\par}
  \vskip 0.2em
  {\Large\bfseries Evidence of an Emergent ``Self'' in Continual Robot Learning\par}
  \vskip 0.4em
  {\normalsize
    \lineskip .25em
    \begin{tabular}[t]{c}
      Adidev Jhunjhunwala, Judah Goldfeder, Hod Lipson
    \end{tabular}\par}
  \vskip 0em
  {\small Creative Machines Lab, Department of Mechanical Engineering, Columbia University, New York, NY\par}
  \vskip 0em
  {\small Creative Machines Lab, Department of Computer Science, Columbia University, New York, NY\par}
\end{center}

\setcounter{section}{0}
\setcounter{subsection}{0}
\renewcommand{\thesection}{S\arabic{section}}
\renewcommand{\thesubsection}{S\arabic{section}.\arabic{subsection}}

% -----------------------------------------------------------------------
\section{Sensitivity of block diagonalization hyperparameters}
\label{app:k_sensitivity}

\begin{figure}[!b]
  \centering
  \includegraphics[width=0.85\textwidth]{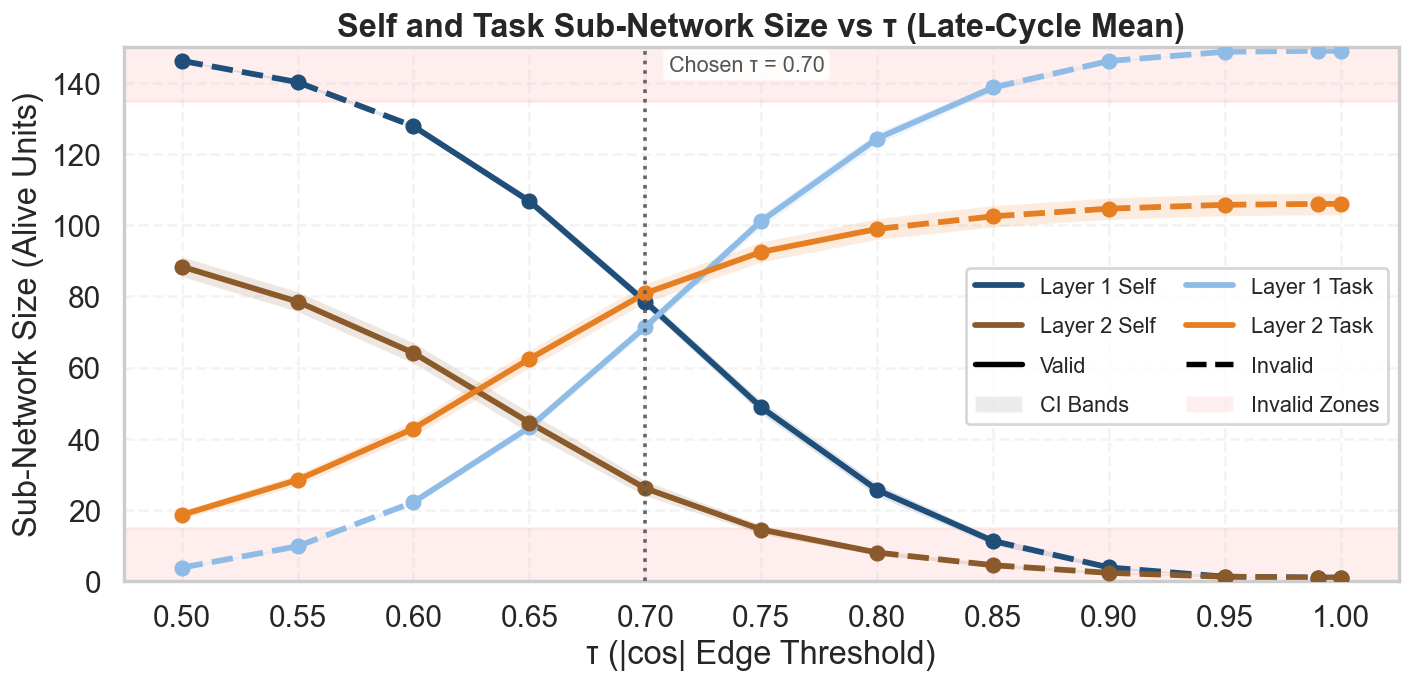}
  \caption{\textbf{Sensitivity to $\tau$ in subnetwork sizes.}
  Late-cycle mean subnetwork sizes (alive units) for the self and task subnetworks in Layers 1--2.
  We selected $\tau=0.70$ as a representative operating point.}
  \label{fig:tau_sensitivity_sizes}
\end{figure}

\begin{figure}[t]
  \centering
  \includegraphics[width=0.85\textwidth]{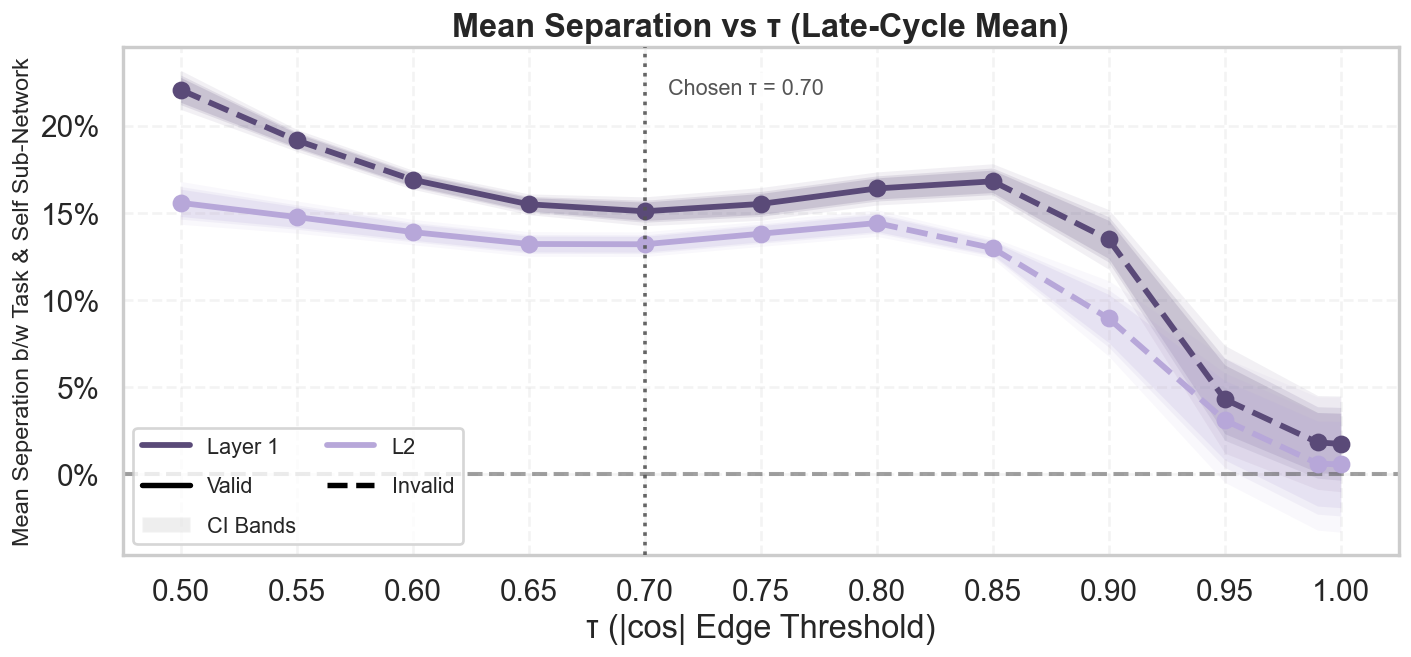}
  \caption{\textbf{Sensitivity to $\tau$ in self--task separation.}
  Late-cycle mean separation $\Delta$ (self minus task persistence score, in percentage points) for Layers 1--2 with confidence bands.
  We selected $\tau=0.70$ as a representative operating point.}
  \label{fig:tau_sensitivity_separation}
\end{figure}

Our block diagonalization procedure had a single hyperparameter: the cosine threshold $\tau$, which determines which edges are retained in the neuron--neuron similarity graph (we keep edges with $|\cos(\cdot,\cdot)| \ge \tau$). Intuitively, lowering $\tau$ produces a denser graph that tends to merge structure into larger subnetworks, while increasing $\tau$ sparsifies the graph and can fragment subnetworks.

We swept $\tau$ over a discrete grid $\{0.50, 0.55, \dots, 0.95, 0.99, 1.00\}$ (12 values) across all $10$ walk-wiggle-bob continual-learning runs, and computed post-stabilisation summaries over cycles $15$--$49$ for both hidden layers. Figure~\ref{fig:tau_sensitivity_sizes} reports the resulting self and task subnetwork sizes, Figure~\ref{fig:tau_sensitivity_separation} reports the mean separation $\Delta$ between self and task persistence scores.

Overall, the key qualitative conclusions were stable for a sensible mid-range of thresholds. In Figure~\ref{fig:tau_sensitivity_sizes}, very low $\tau$ yields an overly dominant self subnetwork (especially in Layer~1), while very high $\tau$ drives the self subnetwork toward near-degenerate size. Over the same range, Figure~\ref{fig:tau_sensitivity_separation} shows that self--task separation remains consistently positive and substantial throughout the mid-range, but collapses rapidly at the most stringent thresholds (e.g.\ $\tau \ge 0.95$), where the decomposition fragments.

We therefore chose $\tau=0.70$ for the main experiments: it lies near the center of the stable region, produces reasonable self and task subnetwork sizes in both layers, and preserves strong, consistent separation across runs.

% -----------------------------------------------------------------------
\section{Intervention analysis}
\label{app:freeze-lesion}

\begin{figure*}[t]
  \centering
  \includegraphics[width=0.5\textwidth]{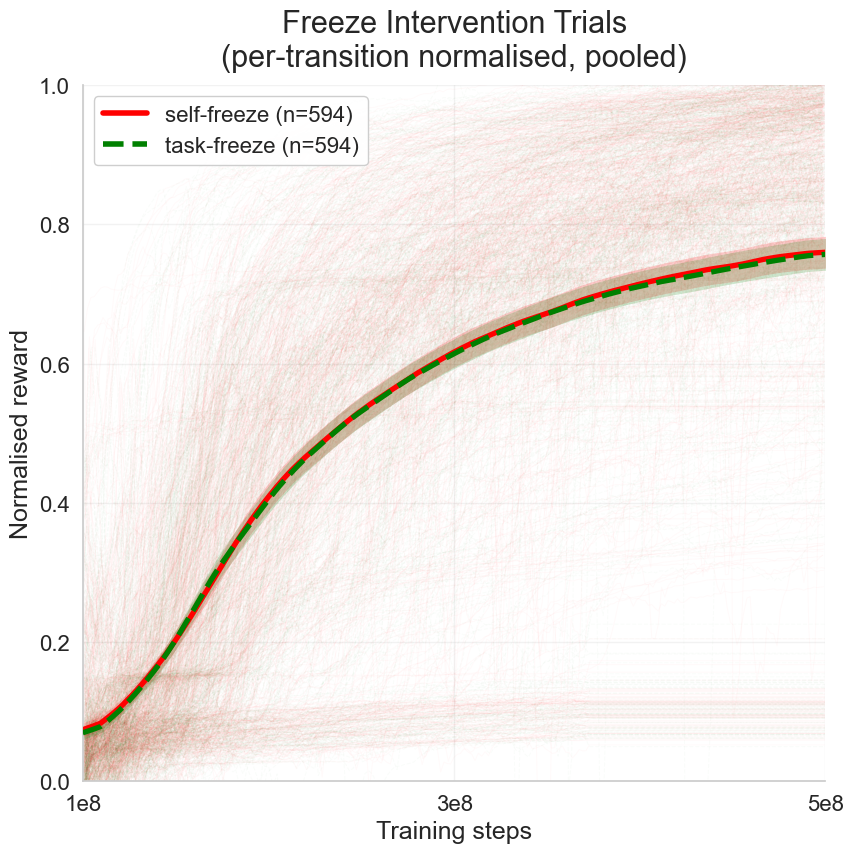}\hfill
  \includegraphics[width=0.5\textwidth]{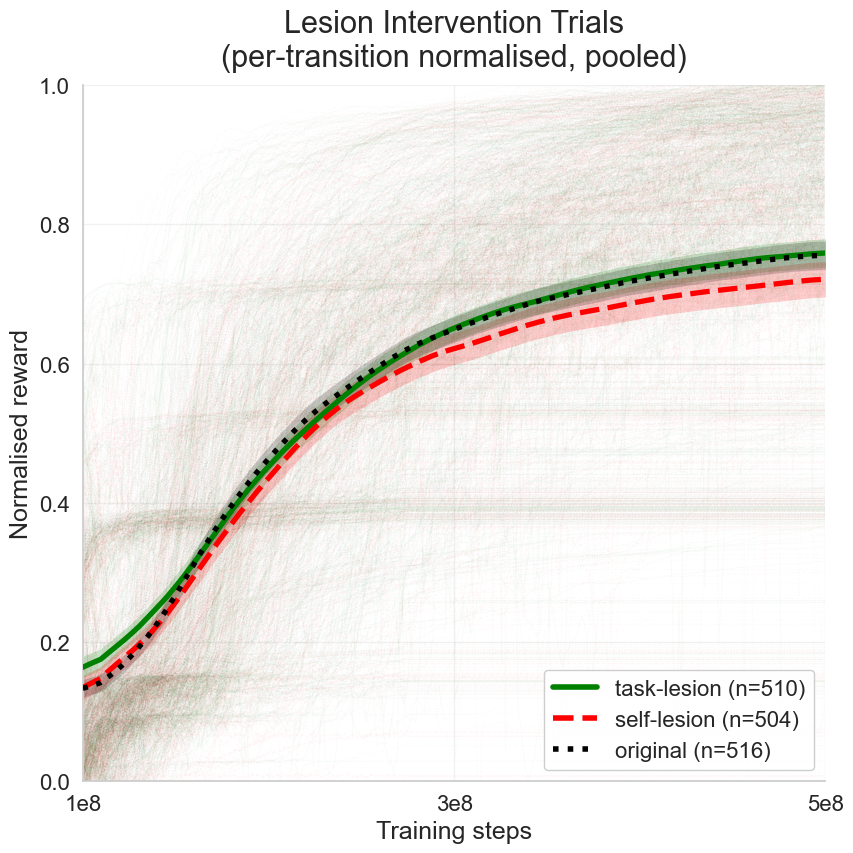}
  \caption{\textbf{Pooled transfer overlays for freeze and lesion interventions.}
  \emph{Left:} per-transition normalized reward curves aggregated across matched self-freeze and task-freeze transfer runs. The self-freeze condition remains slightly above task-freeze across much of the evaluation window. \emph{Right:} per-transition normalized reward curves aggregated across matched self-lesion, task-lesion, and original transfer runs. The task-lesion condition remains above self-lesion over much of the evaluation window. In both panels, faint traces show individual runs and bold curves show the pooled mean trends.}
  \label{fig:freeze-lesion-overlays}
\end{figure*}

Matched transfer runs were compared over the evaluation window from warmup to plateau ($1\times10^8$ to $5\times10^8$ training steps). We report both direct paired reward differences and a within-transition rank-normalized comparison; the latter yielded even stronger support in the predicted direction ($\gg 99\%$ bootstrap support for both interventions).

In both intervention pipelines, we controlled the size of the perturbation so that self and task were compared on equal footing. In the freeze analysis, the number of perturbed neurons in each layer was set to the smaller of the self and task pool sizes, ensuring matched-size interventions. In the lesion analysis, we used a fixed target of 40 neurons per layer, again clipped to the smaller available pool when necessary, and reinitialized the selected neurons while preserving their overall scale statistics, so that the affected portion of the network remained comparable in size and numerical range while its learned function was disrupted. This design prevents differences between self and task conditions from being driven simply by unequal intervention size.

For both analyses, runs were compared in matched pairs sharing the same transition type, random seed, run ID, and source training cycle. In the freeze analysis, across 594 matched transfer pairs (592 non-tied pairs), self-freeze showed a median advantage of 1.6\% of the evaluation window (1.9\% mean). Bootstrap resampling preserved the predicted direction in 92.7\% of resampled comparisons. In the lesion analysis, across 504 matched transfer pairs, task-lesion showed a median advantage of 0.6\% of the evaluation window (2.6\% mean) over self-lesion, and bootstrap resampling preserved the predicted direction in 91.3\% of resampled comparisons.

% -----------------------------------------------------------------------
\section{Hexapod replication}
\label{app:hexapod}

We replicated the walk--wiggle--bob pipeline on a six-legged hexapod with three actuated joints per leg (18 actuated degrees of freedom) across six independent seeds. Both actor and critic were 2-layer MLPs with 250 hidden units per layer and ReLU activations; the main settings followed the quadruped pipeline, with system-specific parameters summarized in Section~\ref{app:rl-details}.

\begin{figure}[t]
  \centering
  \includegraphics[width=\linewidth]{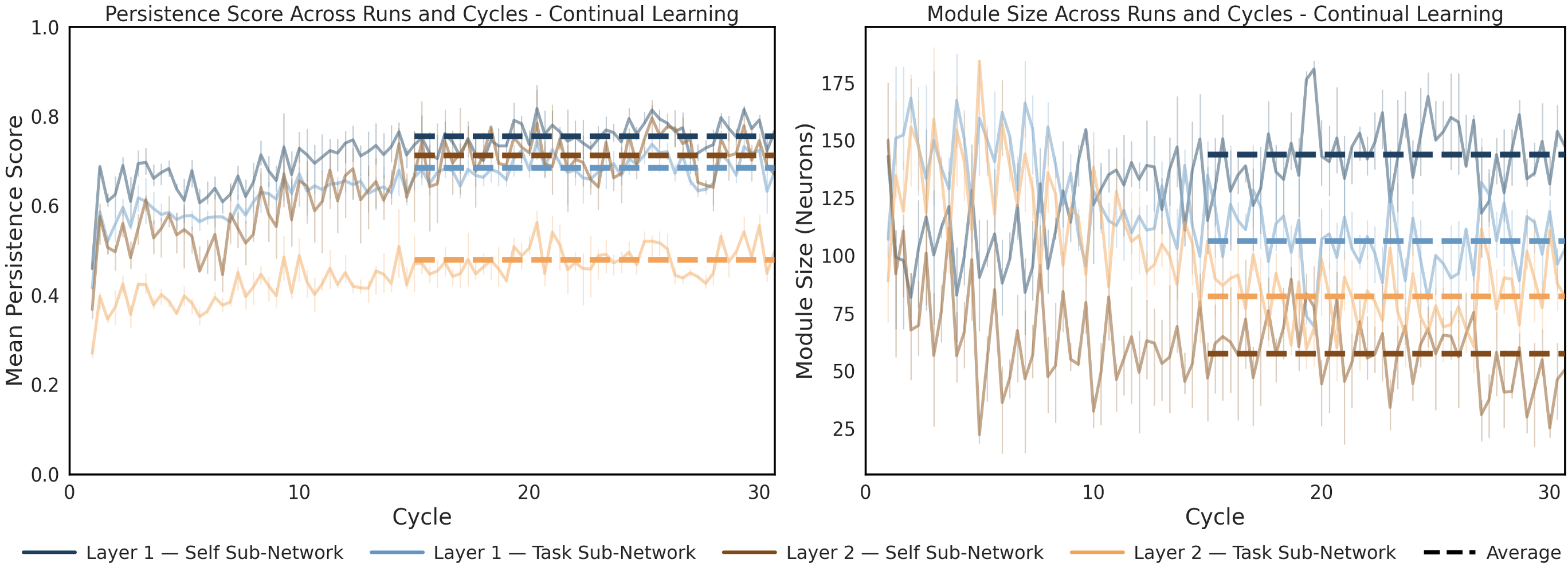}
  \caption{\textbf{Hexapod: subnetwork persistence and size across cycles.}
  Mean persistence score (left) and subnetwork size (right) across 30 training cycles, mirroring the format of Figure~4 in the main text. The self subnetwork consistently exhibits higher mean persistence than the task subnetwork in both layers throughout training.}
  \label{fig:hexa-evolution}
\end{figure}

The self–task separation was consistent. In contrast to the quadruped, the persistence gap was larger in Layer 2, although the self-like subnetwork remained larger in Layer 1.

\subsection{Sensitivity to $\tau$}

\begin{figure}[t]
  \centering
  \includegraphics[width=0.85\textwidth]{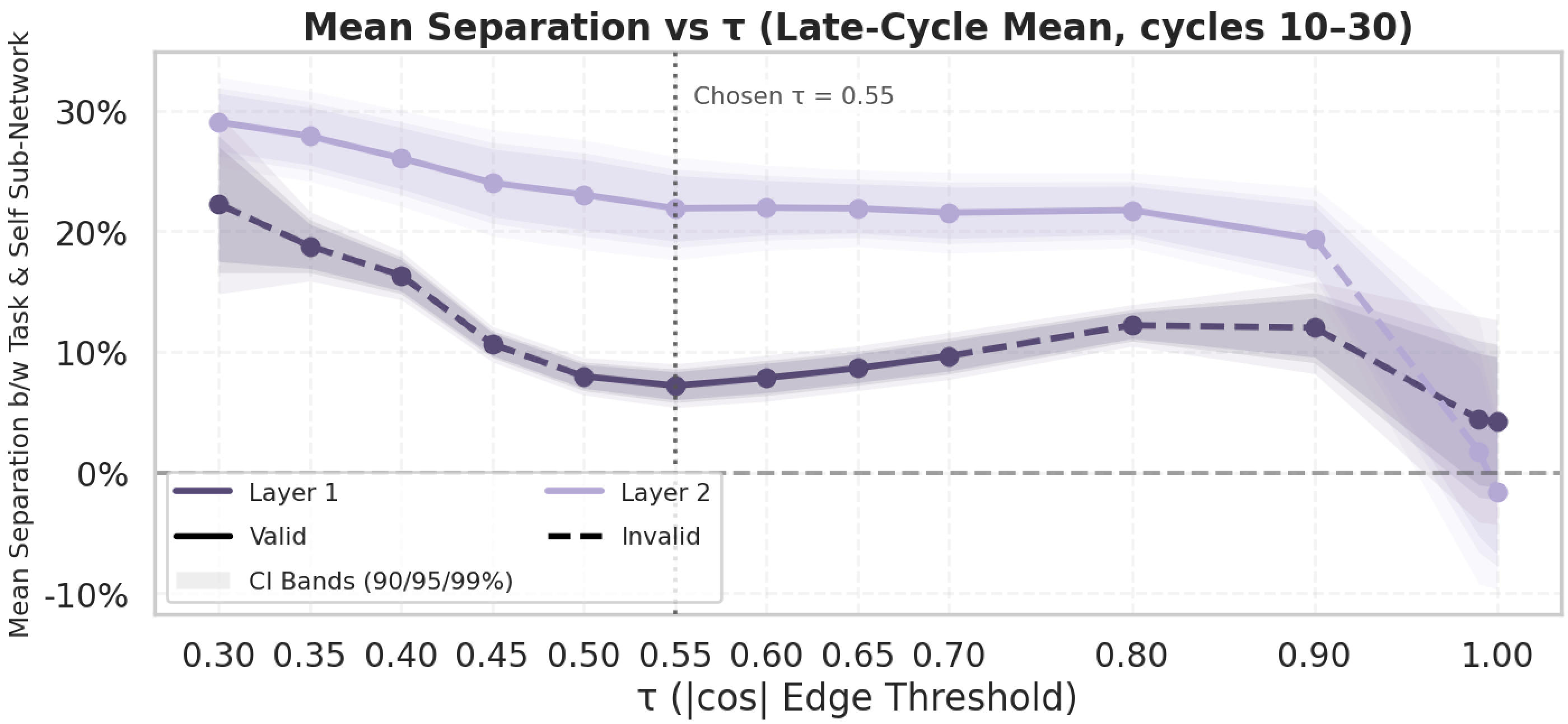}\\[8pt]
  \includegraphics[width=0.85\textwidth]{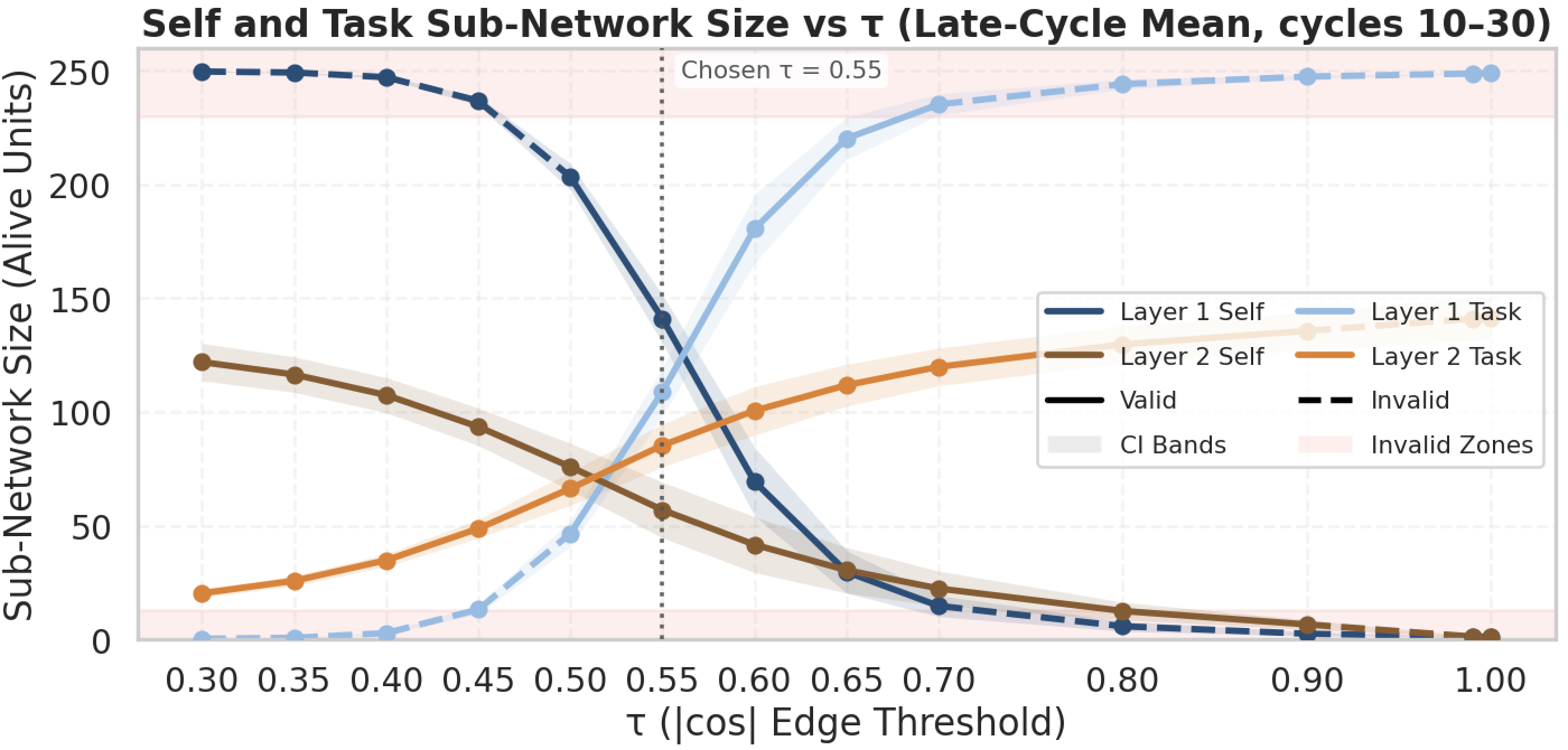}
  \caption{\textbf{Hexapod: sensitivity to $\tau$.}
  \textbf{Top:} Late-cycle mean separation $\Delta$ for Layers 1--2, averaged over cycles 10--30. Solid lines indicate valid decompositions; dashed lines indicate degenerate regimes.
  \textbf{Bottom:} Late-cycle mean subnetwork sizes for Layers 1--2. Shaded invalid zones mark degenerate regimes.}
  \label{fig:hexa-tau}
\end{figure}

The hexapod's wider 250-unit layers required a lower operating threshold ($\tau=0.55$, versus $\tau=0.70$ for the quadruped) to avoid degenerate decompositions; within the valid range, separation remained consistently positive in both layers.

% ----------------------------------------------------------------------

\section{Continual World manipulation results}
\label{app:continual-world-results}

\begin{figure}[t]
    \centering
    \includegraphics[width=\textwidth]{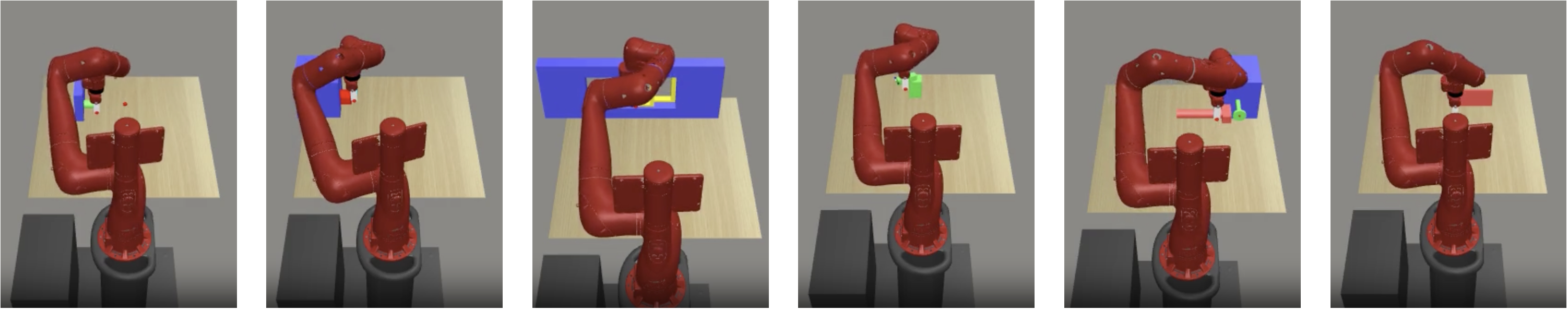}
    \caption{\textbf{Manipulation tasks used in the Continual World replication.} The six Meta-World manipulation tasks used across the four post-validation task sets, spanning tool use, pushing, pressing, sliding, closing, and alignment-sensitive extraction.}
    \label{fig:cw-tasks}
\end{figure}

\begin{figure}[b]
    \centering
    \includegraphics[width=\linewidth]{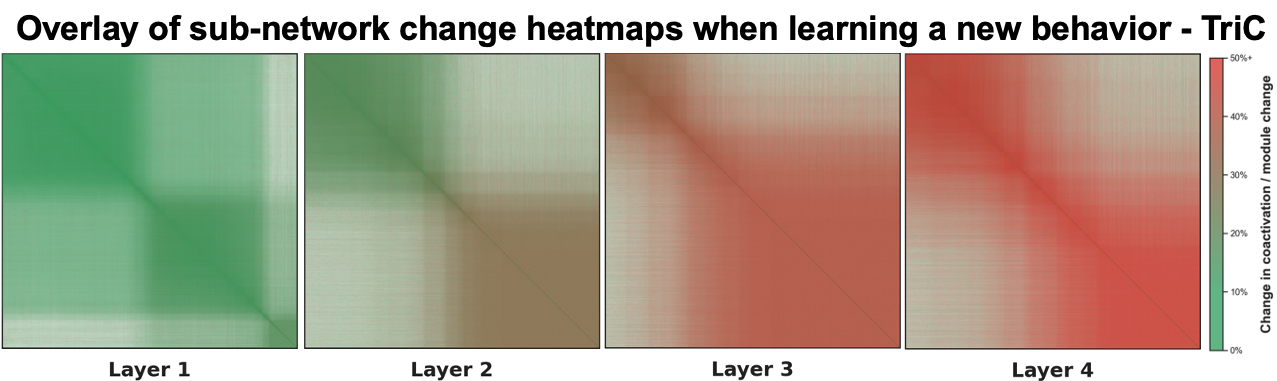}
    \caption{\textbf{triC subnetwork change heatmaps.} The self-like subnetwork changes less than the task-like remainder, with lower relative change in earlier layers.}
    \label{fig:cw-triC-heatmaps}
\end{figure}

We replicated the persistence analysis on a continual manipulation benchmark built on Continual World and Meta-World~\cite{wolczyk2021continualworld,yu2020metaworld}, training a single SAC policy sequentially across ordered task lists with plateau-based phase transitions and applying the same post-hoc analysis as in the primary experiments. The change heatmaps in Figure~\ref{fig:cw-triC-heatmaps} provide a qualitative validation of the same pattern: the self-like subnetwork shows lower relative change than the task-like remainder, with earlier layers generally changing less than later layers.

We constructed four task sets to probe persistence across manipulation behaviors with different contact structure, motion geometry, and control demands. The selected sets span tool use, articulated-object interaction, obstacle-aware pushing, pressing, sliding, and alignment-sensitive extraction. Figure~\ref{fig:cw-tasks} shows the six manipulation tasks used across these sets. Across all four trial families (triA, triC, quadA, and pentaA), the self-like subnetwork remained consistently more persistent than the pooled task subnetwork throughout training, yielding a clear self--task separation in an entirely different control domain.

\subsection{Consistent persistence pattern across manipulation task sets}
\label{app:cw-replication-results}

\begin{figure}[!htbp]
    \centering
    \includegraphics[width=\linewidth]{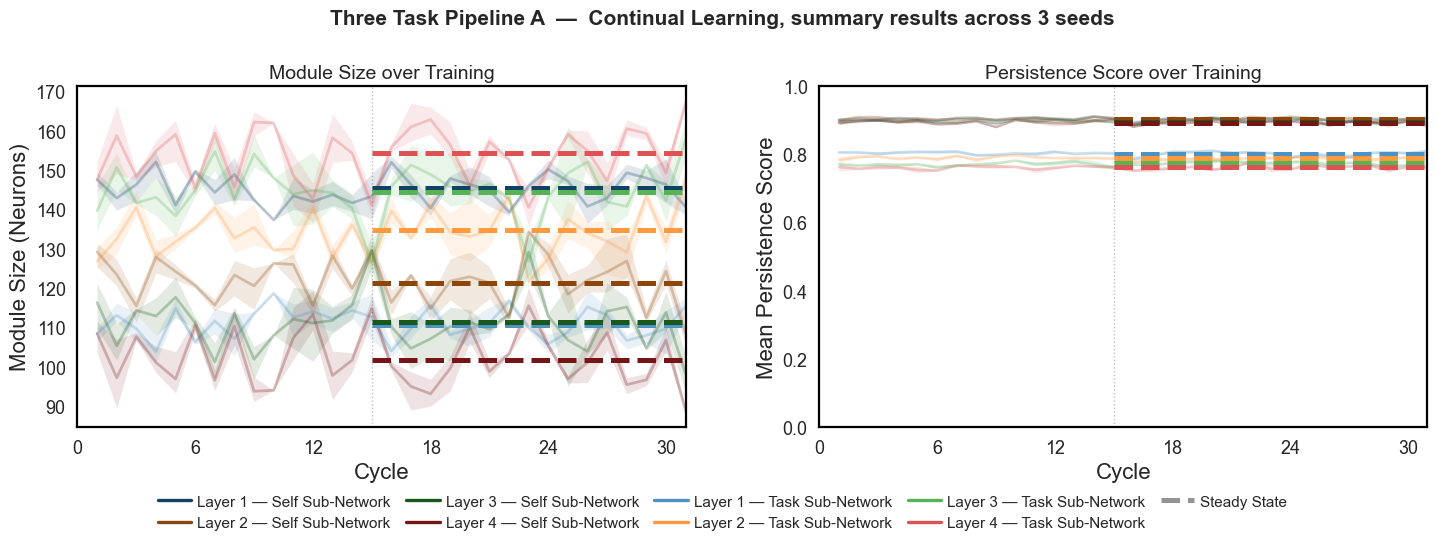}
    \caption{\textbf{triA.} Manipulation results for \texttt{hammer-v1}, \texttt{faucet-close-v1}, and \texttt{peg-unplug-side-v1}.}
    \label{fig:cw-triA}
\end{figure}

\begin{figure}[t]
    \centering
    \includegraphics[width=\linewidth]{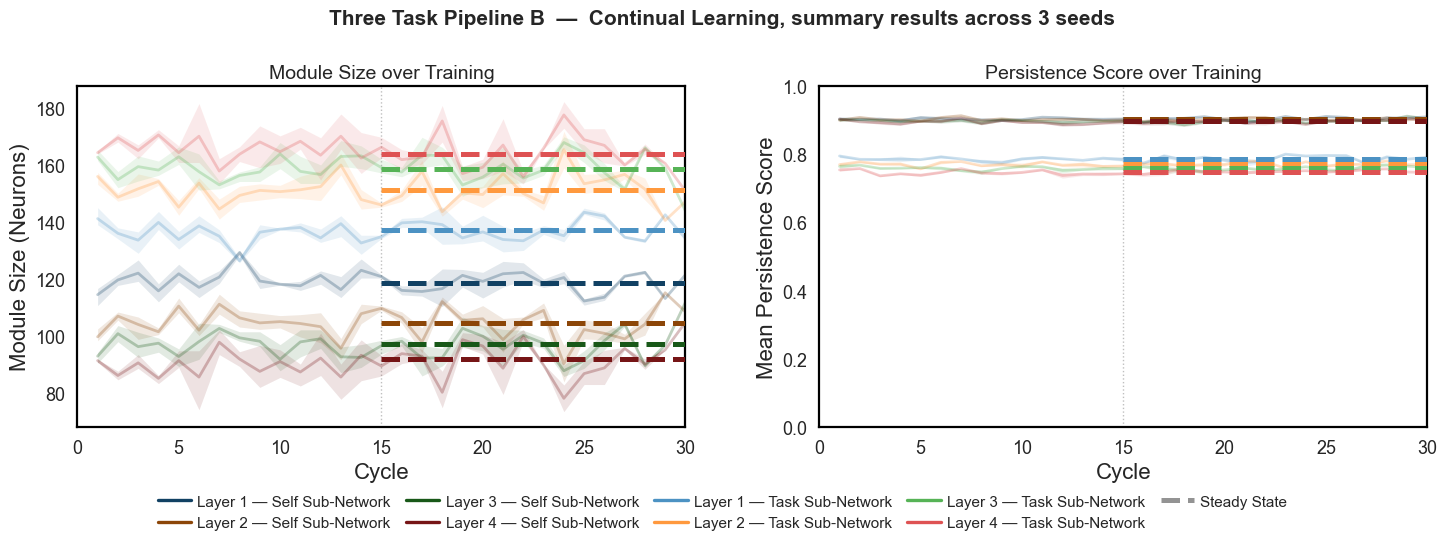}
    \caption{\textbf{triC.} Manipulation results for \texttt{hammer-v1}, \texttt{push-wall-v1}, and \texttt{window-close-v1}.}
    \label{fig:cw-triC}
\end{figure}

\begin{figure}[!htbp]
    \centering
    \includegraphics[width=\linewidth]{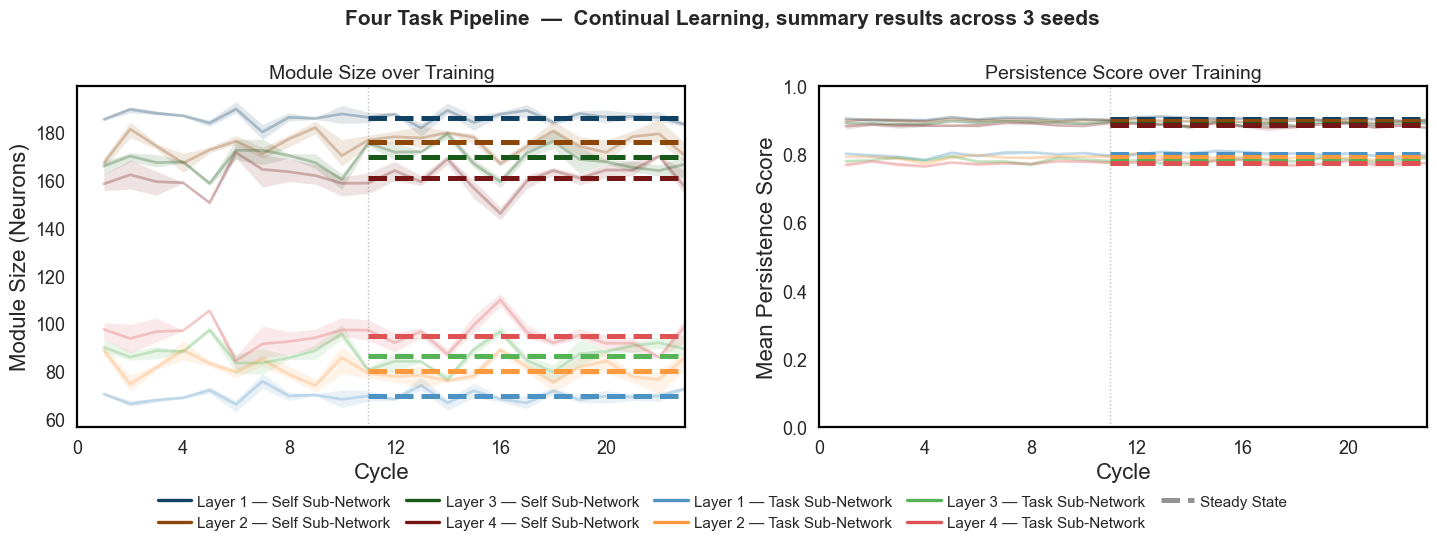}
    \caption{\textbf{quadA.} Manipulation results for \texttt{faucet-close-v1}, \texttt{handle-press-side-v1}, \texttt{window-close-v1}, and \texttt{peg-unplug-side-v1}.}
    \label{fig:cw-quadA}
\end{figure}

\begin{figure}[!htbp]
    \centering
    \includegraphics[width=\linewidth]{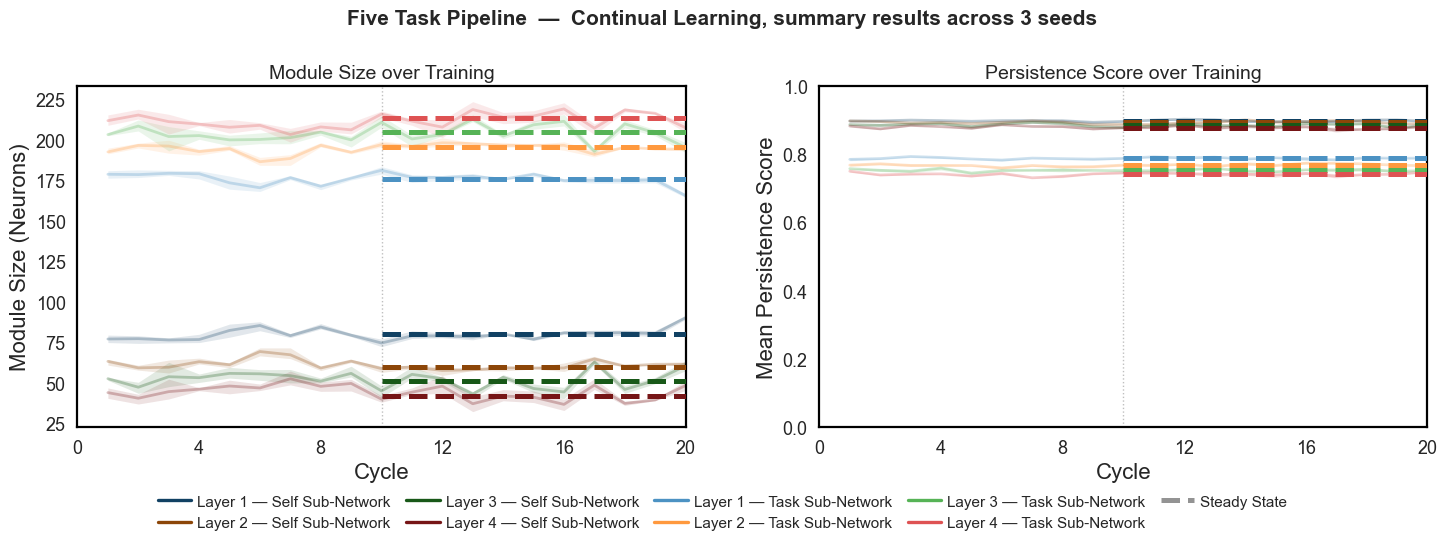}
    \caption{\textbf{pentaA.} Manipulation results for \texttt{hammer-v1}, \texttt{push-wall-v1}, \texttt{faucet-close-v1}, \texttt{stick-pull-v1}, and \texttt{peg-unplug-side-v1}.}
    \label{fig:cw-pentaA}
\end{figure}

Figures~\ref{fig:cw-triA}--\ref{fig:cw-pentaA} summarize the results for the four task sets, each averaged across three seeds. In every case, the largest inferred subnetwork remained more persistent than the pooled task-like remainder, and this separation was sustained across training rather than appearing only at isolated checkpoints. The left panels track inferred module sizes over training and the right panels track mean persistence score. For these trials we used block-diagonalization threshold $\tau=0.9$, and the qualitative separation remained robust. We also observed that the size of the self-like subnetwork varied across task-set breadth, suggesting that the amount of shared persistent structure may depend on how many and what kinds of tasks are included --- an interesting direction for future work.
\FloatBarrier
% -----------------------------------------------------------------------
\section{Aggregated overlay and representative outputs}
\label{app:tessellation-and-examples}

\begin{figure*}
  \centering
  \includegraphics[width=0.85\textwidth]{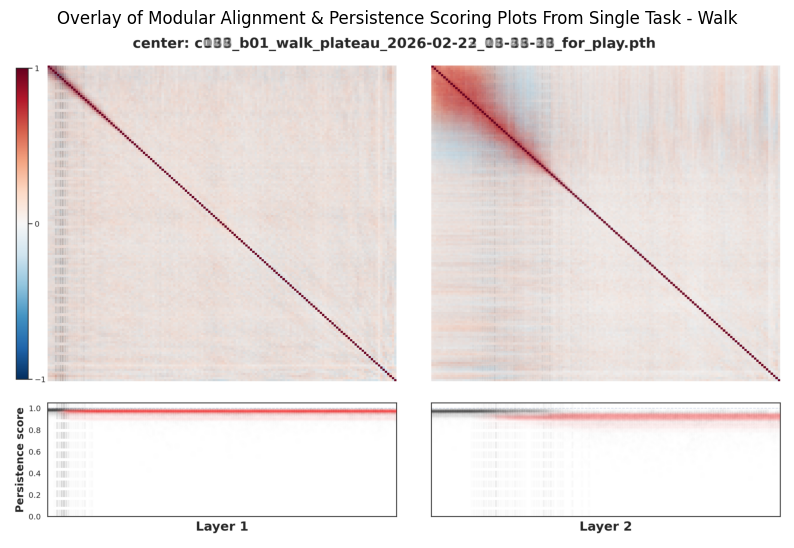}
  \vspace{0.6em}
  \includegraphics[width=0.85\textwidth]{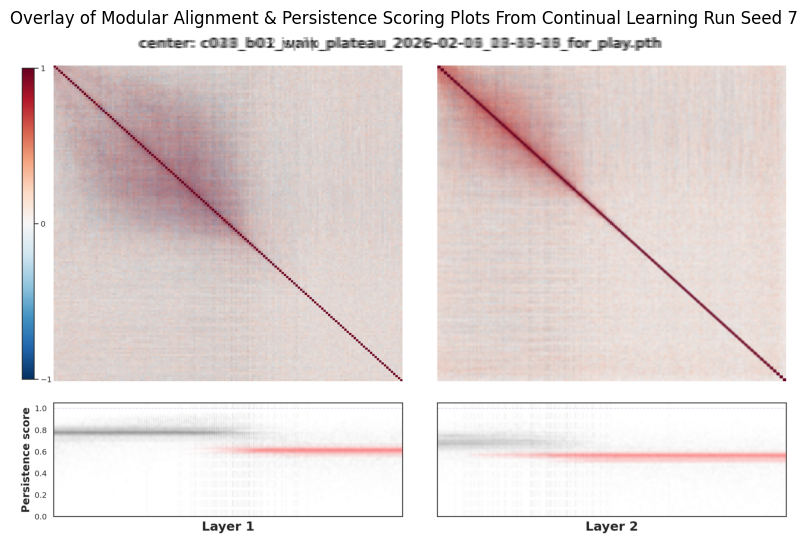}
  \caption{\textbf{Overlay view of activation matrices and persistence-scores.} Each composite is formed by alpha-blending activation-matrix visualizations from all successful plateaued snapshots within a continual-learning run (cycles 16--50), so consistent structure accumulates while inconsistent correlations fade. \textbf{Top:} single-behavior (walk-only) controls remain diffuse and ``spaghetti-like'' with no equally clean core. \textbf{Bottom:} multi-behavior training yields a clear, recurring self-like block in the first hidden layer. In the second hidden layer, both settings show substantial structure, but multi-behavior policies retain a clearer self--task distinction in persistence-scores, while single-behavior controls show a weaker separation.}
  \label{fig:tessellation}
\end{figure*}

\begin{figure*}
  \centering
  \includegraphics[width=0.49\textwidth]{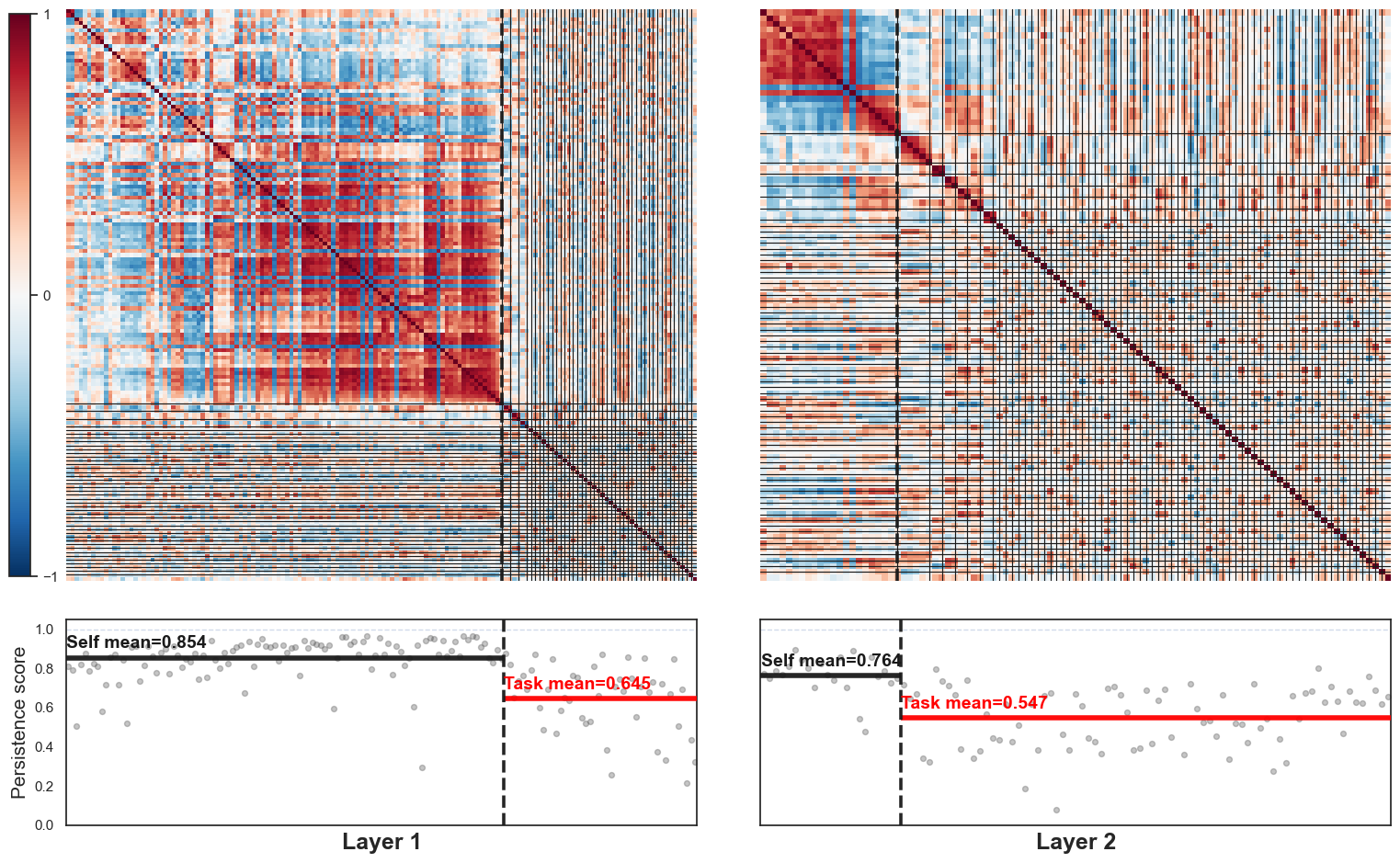}\hfill
  \includegraphics[width=0.49\textwidth]{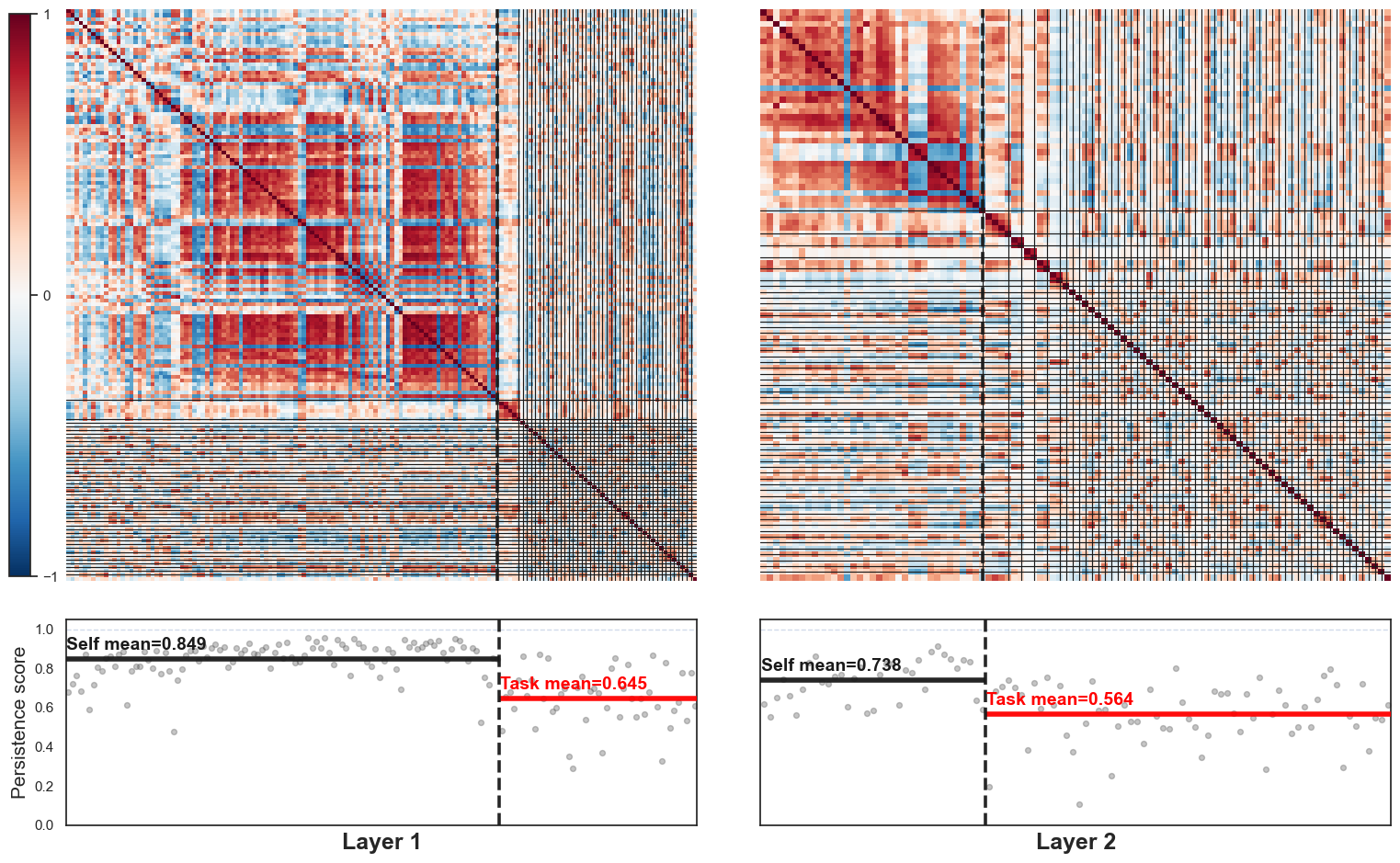}
  \caption{\textbf{Examples of Fig.~3 quantitative outputs (1/3).} Run 2 Cycle 022 (Walk/Wiggle).}
  \label{fig:fig3-selected-examples-1}
\end{figure*}

\begin{figure*}
  \centering
  \includegraphics[width=0.49\textwidth]{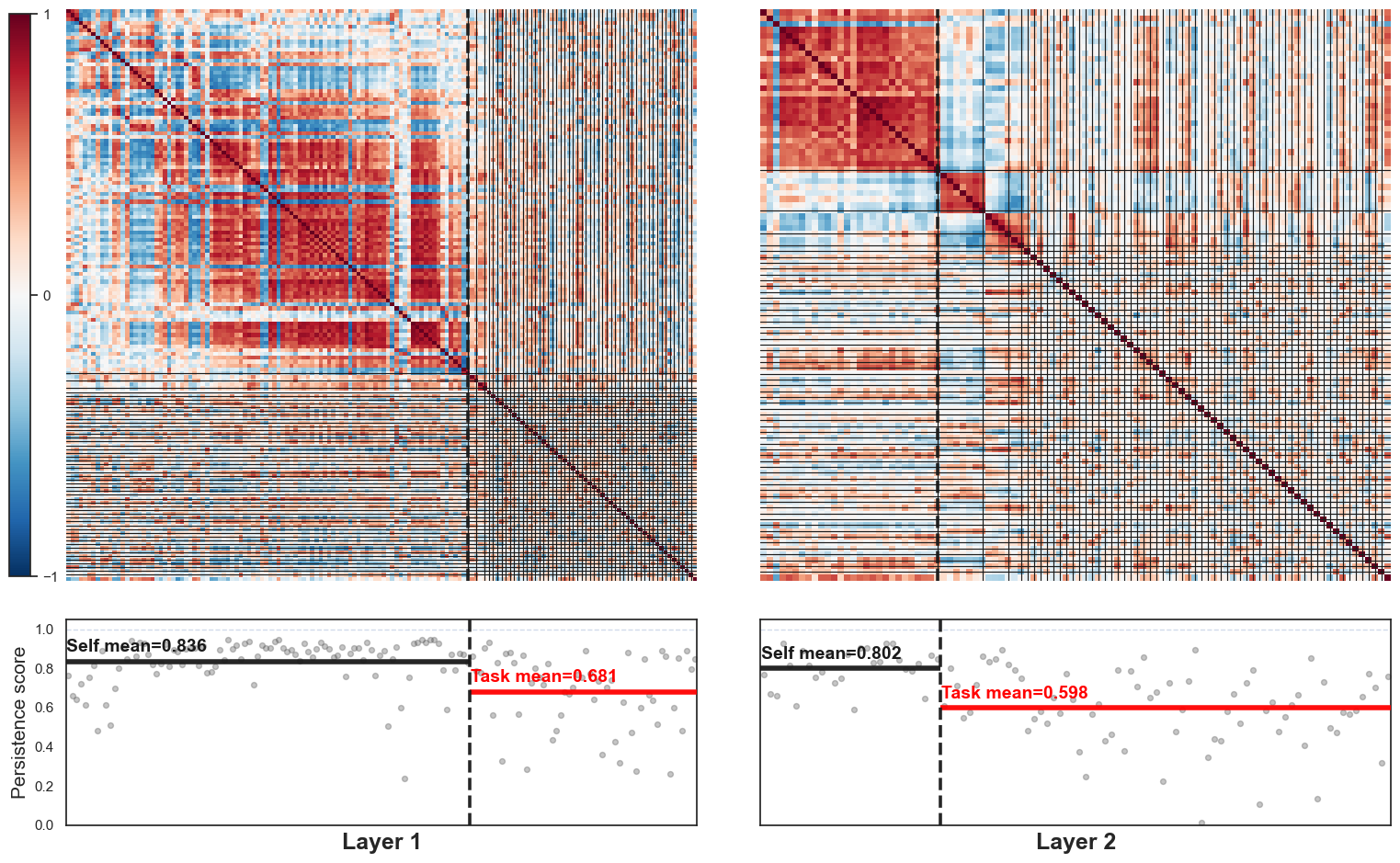}\hfill
  \includegraphics[width=0.49\textwidth]{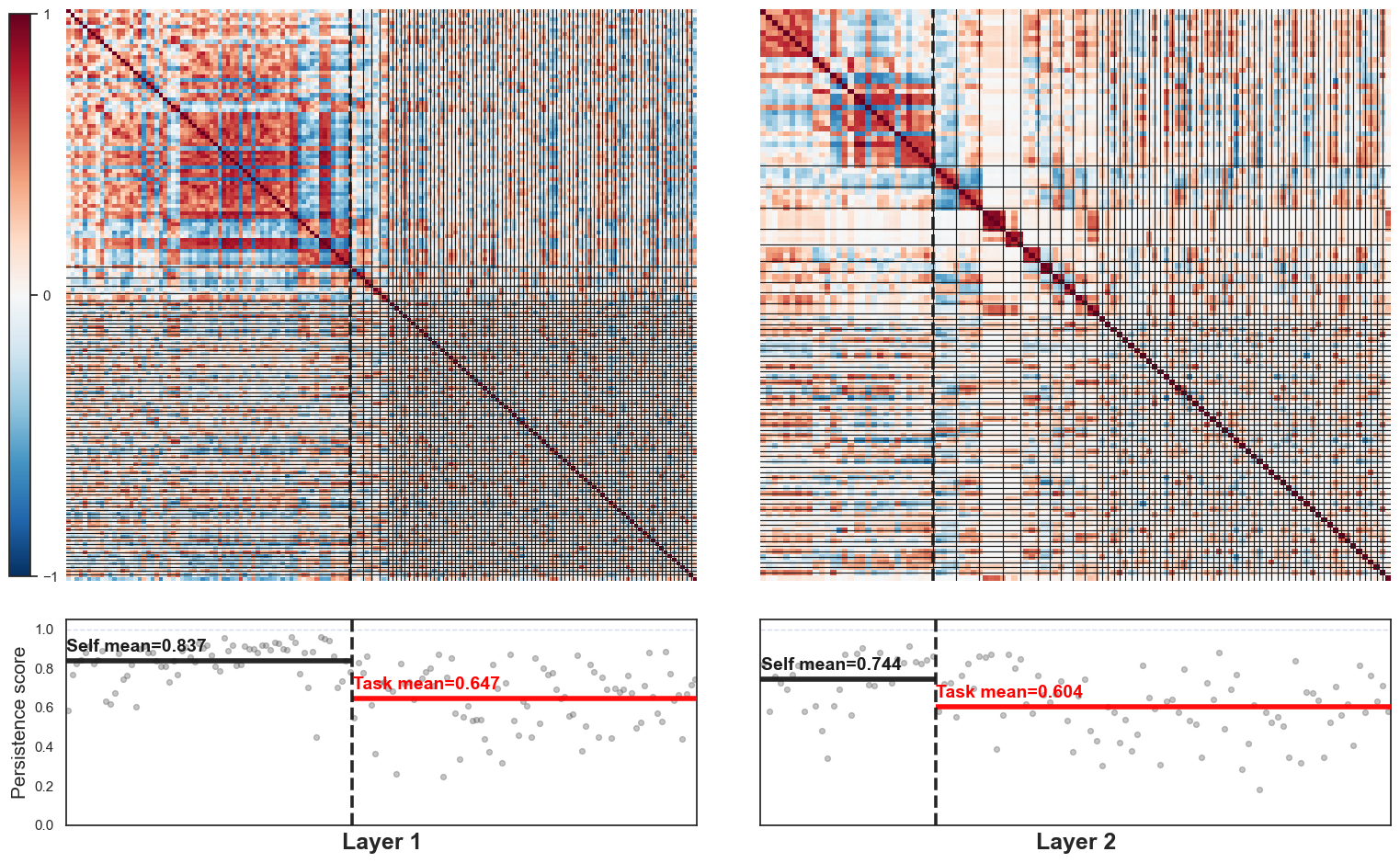}
  \caption{\textbf{Examples of Fig.~3 quantitative outputs (2/3).} Run 2 Cycle 022 (Bob) + Run 7 Cycle 036 (Walk).}
  \label{fig:fig3-selected-examples-2}
\end{figure*}

\begin{figure*}
  \centering
  \includegraphics[width=0.49\textwidth]{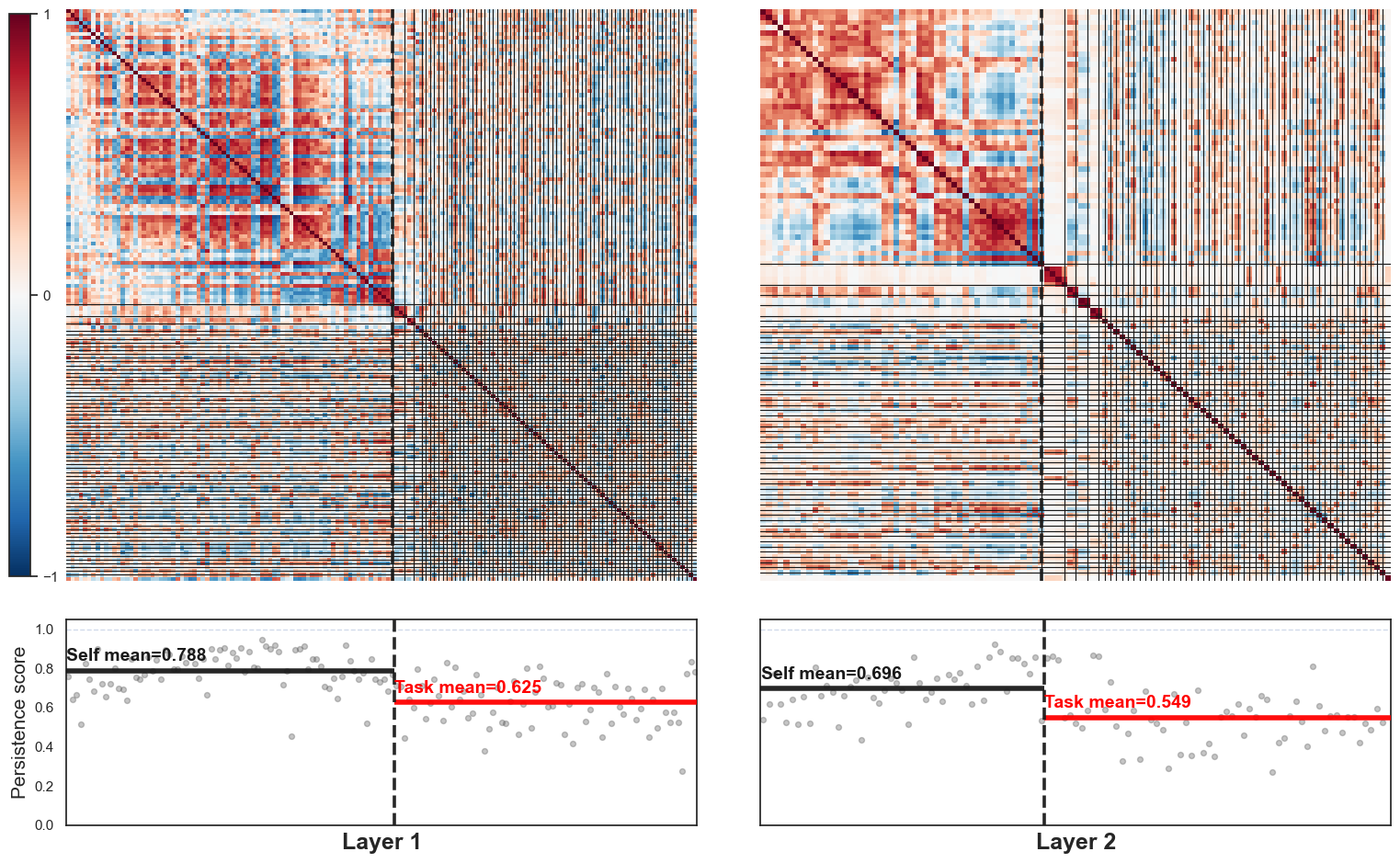}\hfill
  \includegraphics[width=0.49\textwidth]{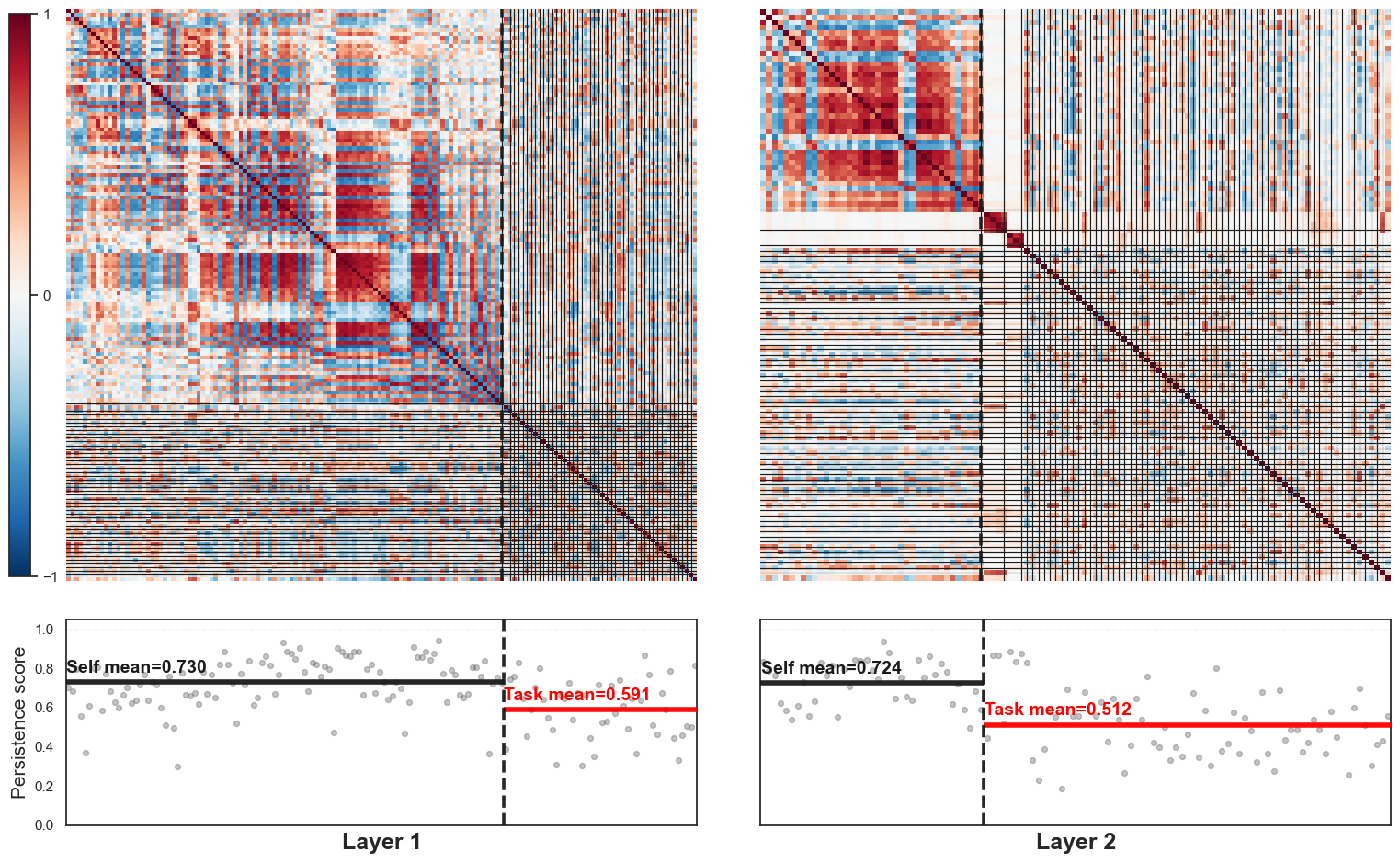}
  \caption{\textbf{Examples of Fig.~3 quantitative outputs (3/3).} Run 7 Cycle 036 (Wiggle/Bob).}
  \label{fig:fig3-selected-examples-3}
\end{figure*}

To show that the modular patterns in the main text recur consistently, we alpha-blended co-activation matrices and persistence-score views from all plateaued snapshots within a run (cycles 16--50) in a shared neuron ordering, so consistent structure accumulates while inconsistent correlations wash out. Single-behavior controls remain diffuse throughout (Fig.~\ref{fig:tessellation}, top), while multi-behavior policies show a clear recurring self-like block in Layer~1 (Fig.~\ref{fig:tessellation}, bottom). Representative per-snapshot outputs across multiple runs, cycles, behaviors, and layers follow (Figs.~\ref{fig:fig3-selected-examples-1}--\ref{fig:fig3-selected-examples-3}); each panel shows the reordered co-activation matrix (top) and per-neuron persistence score (bottom) in the same ordering.

% -----------------------------------------------------------------------
\section{Significance tests for self--task separation}
\label{app:pstats}

For interpretability, we quantified how much the network reorganized at a behavior switch using a transition-level statistic built from neuron matching. Consider a single successful transition (e.g., walk $\rightarrow$ wiggle) with a source policy $S$ and target policy $T$. For each hidden layer, we aligned neurons between $S$ and $T$ by a one-to-one matching computed on a shared reference set of states (Hungarian assignment on cosine similarity of activation traces; see Methods). This defined matched neuron pairs $(i, \pi(i))$ where neuron $i$ in $S$ corresponds to neuron $\pi(i)$ in $T$.

\paragraph{Per-neuron change and subnetwork aggregation.} For each matched pair, we computed a persistence-like similarity $p_i \in [0,1]$ between neuron $i$ (source) and its match $\pi(i)$ (target), and converted it to a \emph{percent change}
\begin{equation}
c_i \;=\; 100\,(1 - p_i).
\end{equation}
We then aggregated these neuron-level changes within the \emph{target-defined} subnetworks from our block diagonalization (self-like = largest subnetwork; task-like = all remaining groupings pooled). Let $\mathcal{S}$ be the set of matched neurons whose \emph{target-side} membership is in the self subnetwork, and $\mathcal{T}$ be the corresponding set for the pooled task region. For this transition, we defined the subnetwork-level changes as averages
\begin{equation}
C_{\text{self}} \;=\; \frac{1}{|\mathcal{S}|}\sum_{i\in \mathcal{S}} c_i, \qquad C_{\text{task}} \;=\; \frac{1}{|\mathcal{T}|}\sum_{i\in \mathcal{T}} c_i.
\end{equation}
Finally, we defined the transition-level \emph{separation} statistic
\begin{equation}
\Delta \;=\; C_{\text{task}} - C_{\text{self}},
\end{equation}
so that $\Delta>0$ means the task-like region reorganizes more than the self-like region at that switch.

\paragraph{Dataset of transitions.} We computed $\Delta$ for every \emph{successful} source$\rightarrow$target transition that satisfied our inclusion criteria (both checkpoints successful; source cycle $>15$ within each run so that post-stabilization dynamics dominated; see Fig.~\ref{fig:delta-overlay}). This yielded $n=916$ transitions, each contributing one scalar $\Delta_j$.

\paragraph{Test A: Null hypothesis of no separation.} We tested whether the mean separation across transitions was positive:
\begin{equation}
H_0:\ \mathbb{E}[\Delta]=0 \qquad\text{vs.}\qquad H_1:\ \mathbb{E}[\Delta]>0.
\end{equation}
Let $\bar{\Delta}$ be the sample mean and $s$ be the sample standard deviation across transitions. The standard error is $\mathrm{SE}=s/\sqrt{n}$, and the (large-$n$) one-sided z-statistic is
\begin{equation}
z \;=\; \frac{\bar{\Delta}}{\mathrm{SE}}.
\end{equation}
In our data, $\bar{\Delta}=16.921$ percentage points, $s=3.093$, $n=916$, so $\mathrm{SE}=0.1022$ and
\begin{equation}
z \;=\; \frac{16.921}{0.1022} \;=\; 165.57.
\end{equation}
The corresponding one-sided p-value is $p=\Pr(Z\ge z)$ with $Z\sim\mathcal{N}(0,1)$. Because $z$ is extremely large, we report the p-value on a log scale:
\begin{equation}
\log_{10} p \;\approx\; -5955.64 \qquad\Rightarrow\qquad p \;\approx\; 10^{-5956}.
\end{equation}
This value should be interpreted as a descriptive measure of the consistency of the observed separation under the transition-level normal approximation, rather than as a literal count of independent experimental trials. Transitions are nested within runs and adjacent cycles are not fully independent; nevertheless, the result indicates that the self--task separation is overwhelmingly positive under this analysis. The extreme magnitude of the value reflects the large mean separation and its high consistency across the $10$ independent robustness runs ($\sim$90 transitions per run, $n=916$ total).

\paragraph{Test B: Separation exceeds a 15-point benchmark.} Repeating the test against a more stringent null ($H_0: \mathbb{E}[\Delta]\le 15$) gives $z_{15} = 18.80$ and $\log_{10} p_{15} \approx -78.4$, again rejecting with overwhelming one-sided significance.

\paragraph{Interpretation.} Together, these tests showed that the task-like region reorganized substantially more than the self-like region at behavior switches: the mean separation was $16.92$ percentage points, with a 99\% one-sided lower bound of $16.68$ percentage points, and both the null of no separation and the stronger 15-point benchmark were rejected with overwhelming one-sided significance.

% -----------------------------------------------------------------------
\section{Single-behavior baselines across cycles}
\label{supp:phaseplots-baselines}

As a control, we computed the same subnetwork diagnostics for single-behavior baselines (bob-only and wiggle-only; walk-only is shown in the main text) trained under the same phase structure. Across these baselines, the first hidden layer typically showed little to no separation between the persistence of the self and task subnetworks, consistent with the absence of a cleanly isolatable ``self-like'' core when the objective did not change. In the second hidden layer, a minor separation could emerge in some cases, similar to the walk-only baseline, but the degree of separation was substantially smaller than in the continual-learning checkpoints.

\begin{figure}[p]
  \centering
  \includegraphics[width=0.49\columnwidth]{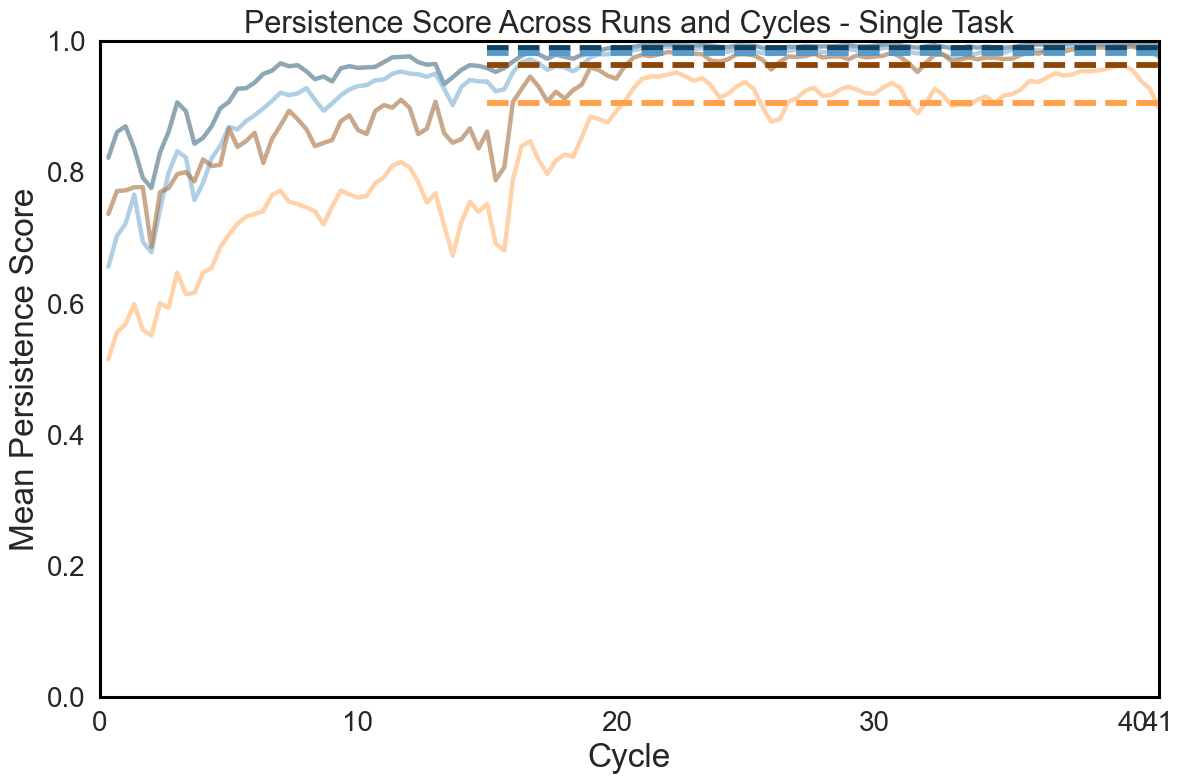}\hfill
  \includegraphics[width=0.49\columnwidth]{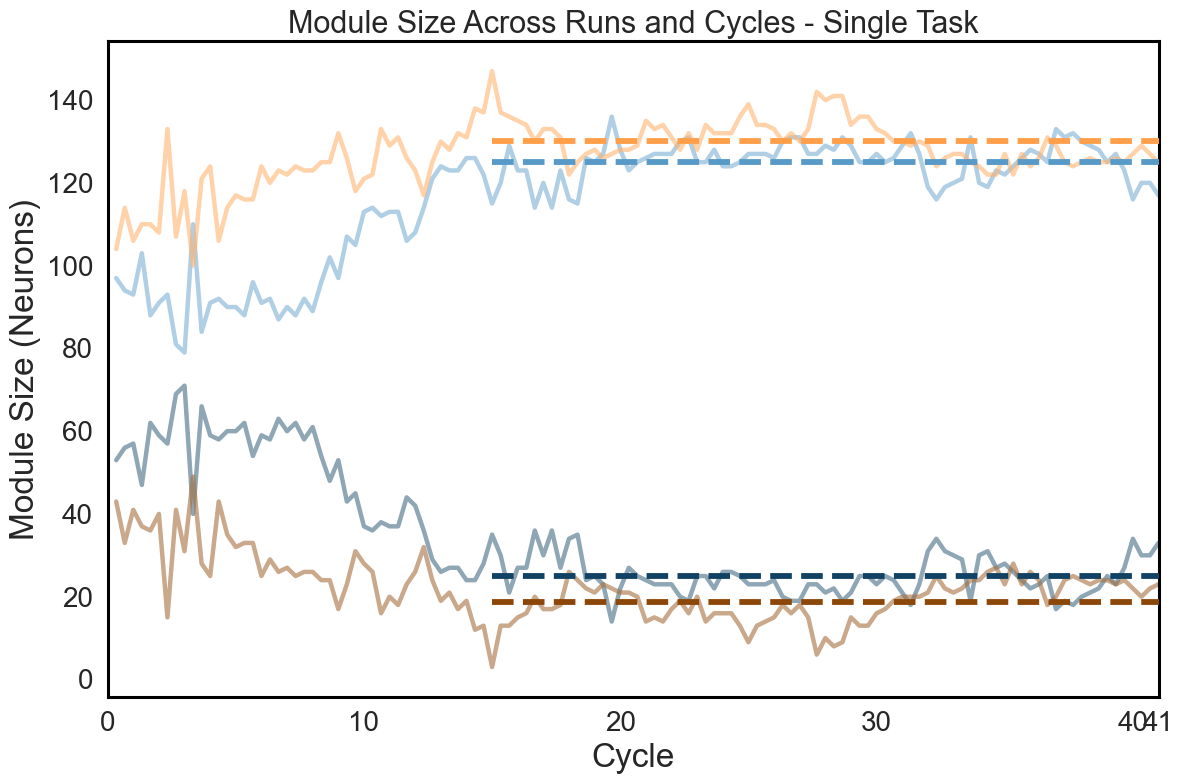}\\[8pt]
  \includegraphics[width=0.49\columnwidth]{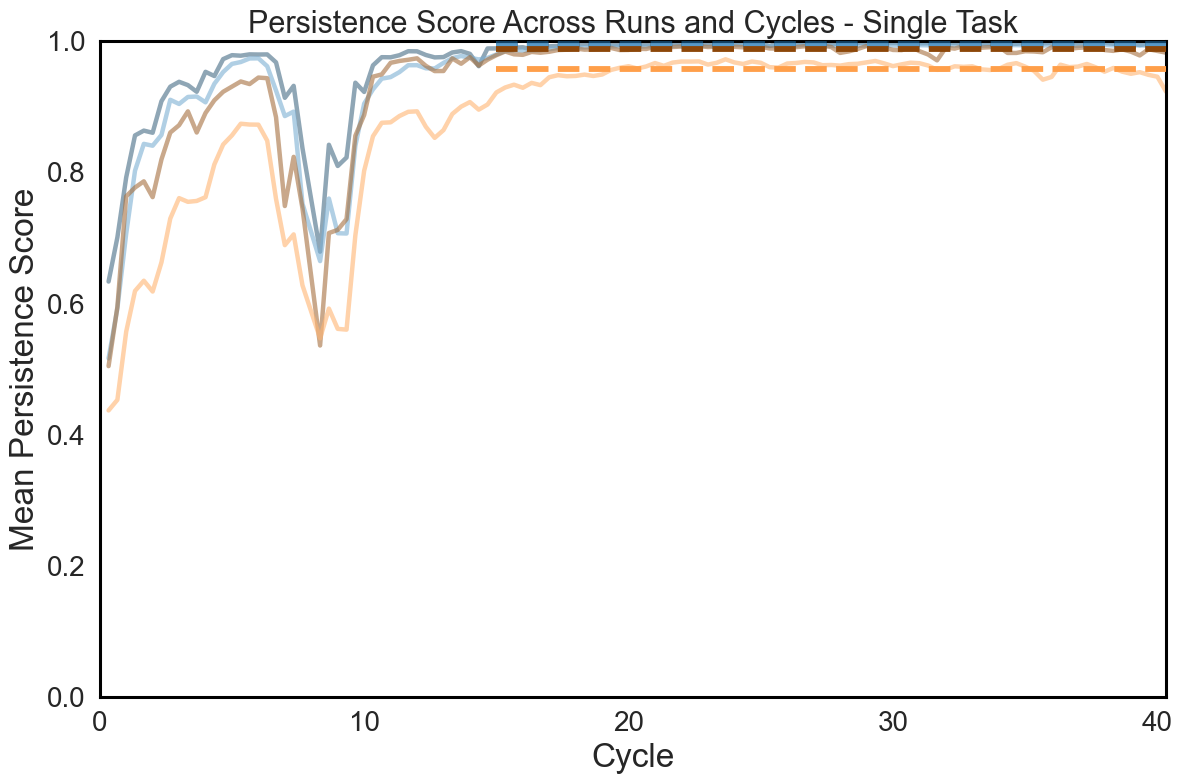}\hfill
  \includegraphics[width=0.49\columnwidth]{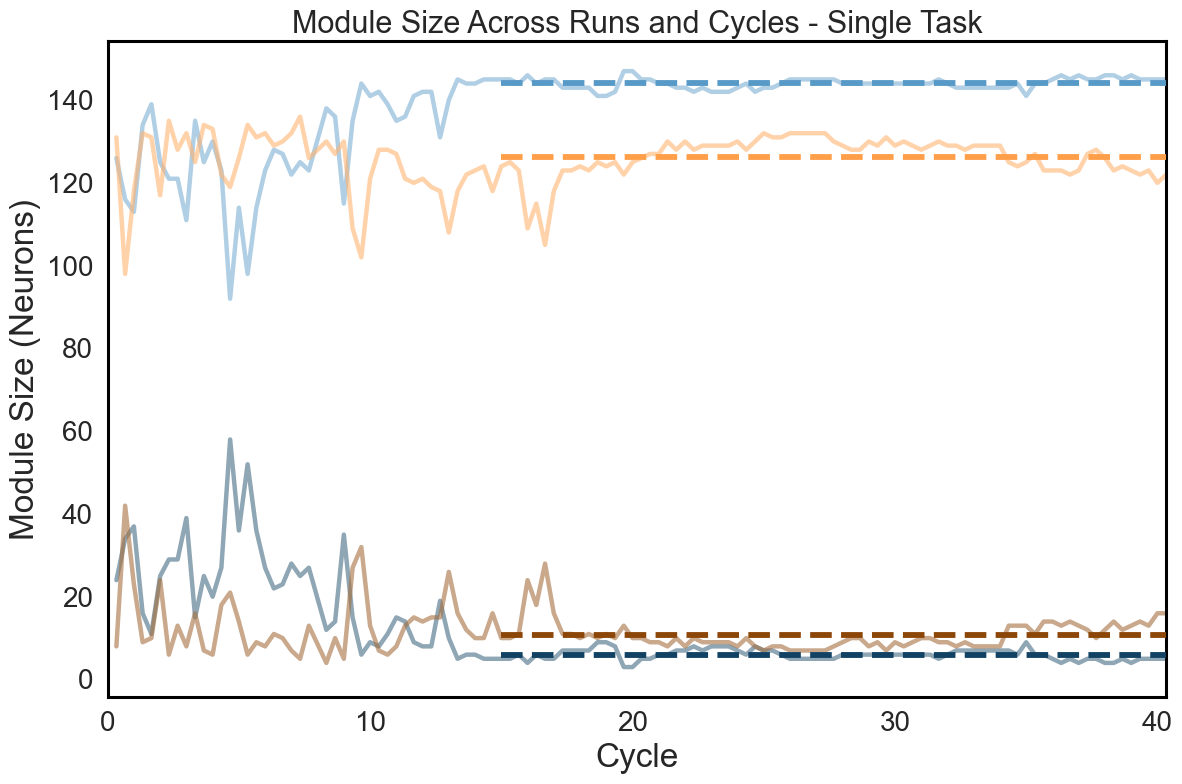}
  \caption{\textbf{Single-behavior baselines: bob-only (top row) and wiggle-only (bottom row).} \emph{Left column:} Persistence score of the self and task subnetworks across phases. \emph{Right column:} Size of the self and task subnetworks across phases.}
  \label{fig:supp-baseline-phaseplots}
\end{figure}

\FloatBarrier

% -----------------------------------------------------------------------
\section{RL hyperparameters across systems}
\label{app:rl-details}

Table~\ref{tab:rl-hyperparams} summarizes the main training, architecture, and environment settings used for the quadruped, hexapod, and manipulation experiments.

\vspace{0.35em}
\begin{center}
\captionof{table}{\textbf{RL hyperparameters across the three experimental systems.}}
\label{tab:rl-hyperparams}
\renewcommand{\arraystretch}{1.03}
\setlength{\tabcolsep}{3.0pt}
\scriptsize
\begin{tabular*}{0.98\textwidth}{@{\extracolsep{\fill}}p{0.32\textwidth}ccc@{}}
\toprule
\textbf{Hyperparameter} & \textbf{Quadruped} & \textbf{Hexapod} & \textbf{CW / Meta-World} \\
\midrule

\multicolumn{4}{l}{\textit{Algorithm and implementation}} \\
Algorithm & \multicolumn{2}{c}{SAC (\texttt{rl\_games}, PyTorch)} & SAC (custom TF) \\
Simulator & \multicolumn{2}{c}{IsaacLab} & Continual World / Meta-World \\
\midrule

\multicolumn{4}{l}{\textit{Network architecture}} \\
Hidden layers & 2 & 2 & 4 \\
Units / layer & 150 & 250 & 256 \\
Activation & ReLU & ReLU & Leaky ReLU \\
First-layer norm & \multicolumn{2}{c}{---} & LayerNorm $\to$ Tanh \\
Obs.\ dim & 48 & 66 & 12 \\
Action dim & 8 & 18 & 4 \\
\midrule

\multicolumn{4}{l}{\textit{SAC core}} \\
Discount $\gamma$ & \multicolumn{3}{c}{$0.99$} \\
Actor / critic lr & \multicolumn{2}{c}{$3\times10^{-4}$} & $1\times10^{-3}$ \\
Temp.\ lr $\alpha_{\text{lr}}$ & \multicolumn{2}{c}{$3\times10^{-3}$} & automatic \\
Target entropy & $-8.0$ & $-6.0$ & automatic \\
Target-net $\tau$ & \multicolumn{2}{c}{$0.003$} & $0.005$ \\
Grad.\ clip & \multicolumn{2}{c}{$1.0$} & --- \\
\midrule

\multicolumn{4}{l}{\textit{Replay and updates}} \\
Batch size & \multicolumn{2}{c}{32\,768} & 256 \\
Replay buffer / phase & \multicolumn{2}{c}{$1\times10^{6}$} & $1\times10^{6}$ \\
Updates / env step & \multicolumn{2}{c}{32} & 0.2 \\
Warmup steps & \multicolumn{2}{c}{10\,000} & 1\,000 \\
\midrule

\multicolumn{4}{l}{\textit{Environment and curriculum}} \\
Parallel envs & 8\,192 & 16\,384 & 8 \\
Episode length & \multicolumn{2}{c}{1\,000} & 200 \\
Steps / phase & \multicolumn{2}{c}{plateau-based} & $1\times10^{6}$ fixed \\
Behaviors / tasks & walk, wiggle, bob & walk, wiggle, bob & manipulation tasks \\
Cycles & 50 & 30 & task-set dependent (100 phases) \\
Seeding & \multicolumn{2}{c}{deterministic phase offset} & fixed per rep. \\
\bottomrule
\end{tabular*}
\end{center}
\vspace{-0.5em}

% -----------------------------------------------------------------------
\section{Plateau detection and Example Rollout}
\label{app:plateau}

We switched behaviors (walk $\rightarrow$ wiggle $\rightarrow$ bob, repeating) using a simple plateau-stop rule defined on episode returns, where episode return is the sum of rewards over an episode. Plateau checks operated on \emph{aggregated environment steps}, i.e., steps summed across all parallel environments, and fixed-size episode windows.

\vspace{0.25em}
\noindent\textbf{Phase stop criteria (per phase for quadruped):}
\begin{itemize}
  \item \textbf{Warm-up guard:} plateau checks are disabled until the phase has seen at least \par \texttt{--plateau\_min\_steps} $= 250{,}000{,}000$ aggregated env-steps.
  \item \textbf{Hard cap:} the phase stops once it reaches \texttt{--max\_steps\_phase} $= 1{,}500{,}000{,}000$ aggregated env-steps, regardless of returns.
  \item \textbf{Plateau decision (episode-based):} after the warm-up guard is satisfied, the phase stops when two consecutive episode-return windows of size \texttt{--plateau\_episode\_window} (default $50{,}000$ episodes) satisfy all of:
  \begin{itemize}
    \item \textbf{Minimum performance:} $\mu_{\text{recent}} \ge \texttt{--plateau\_min\_return}$ (default $500$).
    \item \textbf{Stable mean (relative):} $\frac{|\mu_{\text{recent}}-\mu_{\text{prev}}|}{|\mu_{\text{prev}}|+\varepsilon} \le \texttt{--plateau\_rel\_change}$ (default $0.05$).
    \item \textbf{Std safeguard (absolute):} $|\mu_{\text{recent}}-\mu_{\text{prev}}| \le \texttt{--plateau\_std\_coeff}\cdot\sigma$ over the combined two windows (default $1.0\cdot\sigma$), where $\sigma$ is the return standard deviation over both windows.
  \end{itemize}
\end{itemize}

\begin{figure*}[t]
  \centering
  \includegraphics[width=0.95\textwidth]{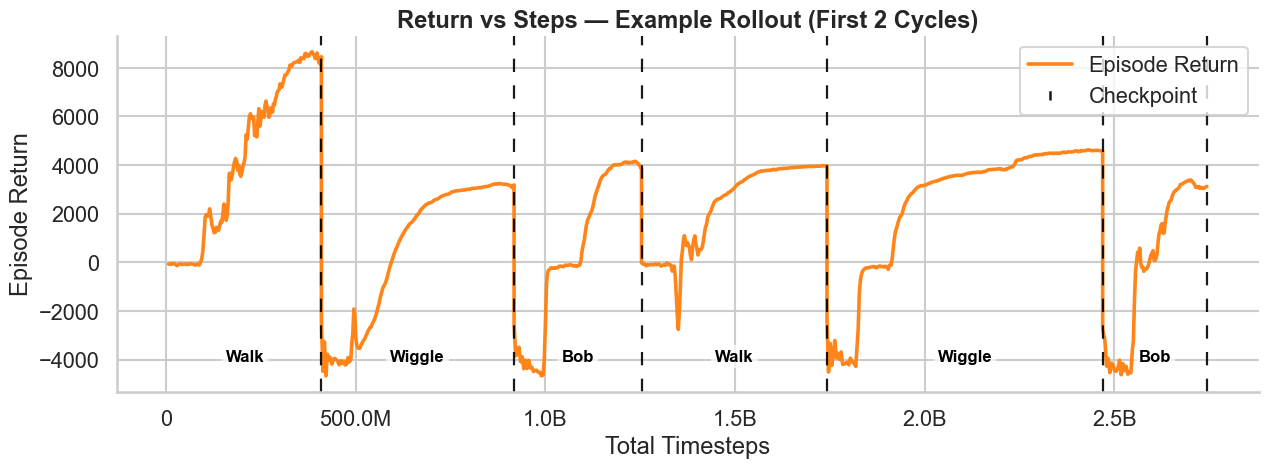}
  \caption{\textbf{Example rollout of the training schedule.} An example training trace illustrating phase switching across two full walk$\rightarrow$wiggle$\rightarrow$bob cycles with the plateau-stop logic described above.}
  \label{fig:plateau-example-rollout}
\end{figure*}

% -----------------------------------------------------------------------
\section{Behavior definitions and rewards}
\label{app:behaviors}

All three behaviors shared the same dynamics, observation, and action spaces; they differed only in their reward shaping.

\paragraph{Walk.} For walking, we used IsaacLab's default Ant locomotion reward, which encourages forward velocity along world $+x$, maintaining an upright posture at a nominal body height, and penalizes large or rapidly changing torques (energy cost). We did not modify this reward; it served as a standard baseline locomotion objective.

\paragraph{Wiggle.} For wiggle, we rewarded rotation about the vertical axis while discouraging lateral drift and jerky actions. Let $s \in \{+1, -1\}$ be the chosen wiggle direction, $\omega_z$ the body-frame yaw rate, $\mathbf{v}_{b,xy}$ the horizontal body-frame velocity, and $\Delta \mathbf{a}$ the change in action between consecutive steps. The wiggle reward is
\begin{align}
  r_{\text{wiggle}} =\ & \alpha \, \max(s \, \omega_z, 0) - \lambda_{\text{back}} \, \max(-s \, \omega_z, 0) \nonumber - \lambda_v \, \lVert \mathbf{v}_{b,xy} \rVert - \lambda_{\text{jerk}} \, \lVert \Delta \mathbf{a} \rVert^2 + k \, \text{streak}(t),
\end{align}
where $\alpha$ sets the gain for wiggling in the desired direction, $\lambda_{\text{back}}$ penalizes wiggling backwards, $\lambda_v$ penalizes drift, $\lambda_{\text{jerk}}$ penalizes jerky torques, and $\text{streak}(t)$ is a small bonus that increases with the duration of a clean wiggle.

\paragraph{Bob.} For bob, we rewarded upward center-of-mass velocity and penalized sideways drift and jerky torques. Let $v_z$ be the vertical velocity of the base (world frame) and $\mathbf{v}_{b,xy}$ the horizontal body-frame velocity. The bob reward is
\begin{align}
  r_{\text{bob}} =\ & \beta \, \max(v_z, 0) - \lambda_{\text{drift}} \, \lVert \mathbf{v}_{b,xy} \rVert \nonumber - \lambda_{\text{jerk}} \, \lVert \Delta \mathbf{a} \rVert^2,
\end{align}
where $\beta$ scales the reward for upward motion, $\lambda_{\text{drift}}$ discourages lateral drift, and $\lambda_{\text{jerk}}$ penalizes abrupt torque changes.

% -----------------------------------------------------------------------
\section{Sim-to-Real Details}
\label{app:readyant}

The MuJoCo Ant and its variants in Gymnasium and Isaac Lab are widely used benchmarks for continuous-control locomotion~\cite{gymnasium_ant,mittal2023isaaclab}. For the sim-to-real demonstrations in this paper, we built a small, low-cost Ant-like quadruped whose joint layout and overall proportions closely follow the canonical Ant morphology. This hardware was designed specifically for \emph{retroactive} sim-to-real: policies were trained in simulation first, and the physical platform was then assembled to replay trained joint trajectories with minimal additional tuning.

\subsection{Design goals}

\begin{figure}[b]
\centering
\includegraphics[width=0.49\columnwidth]{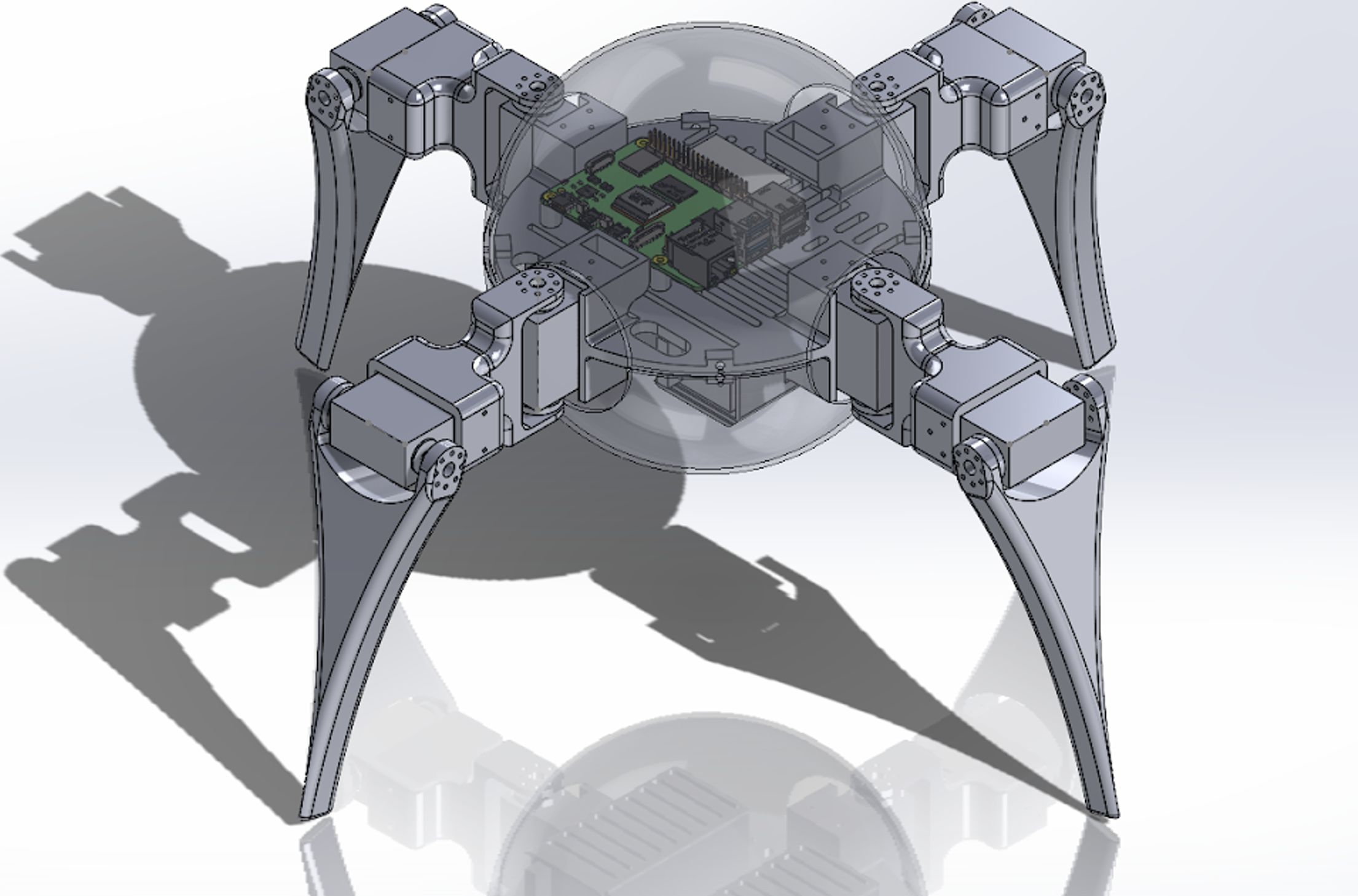}\hfill
\includegraphics[width=0.49\columnwidth]{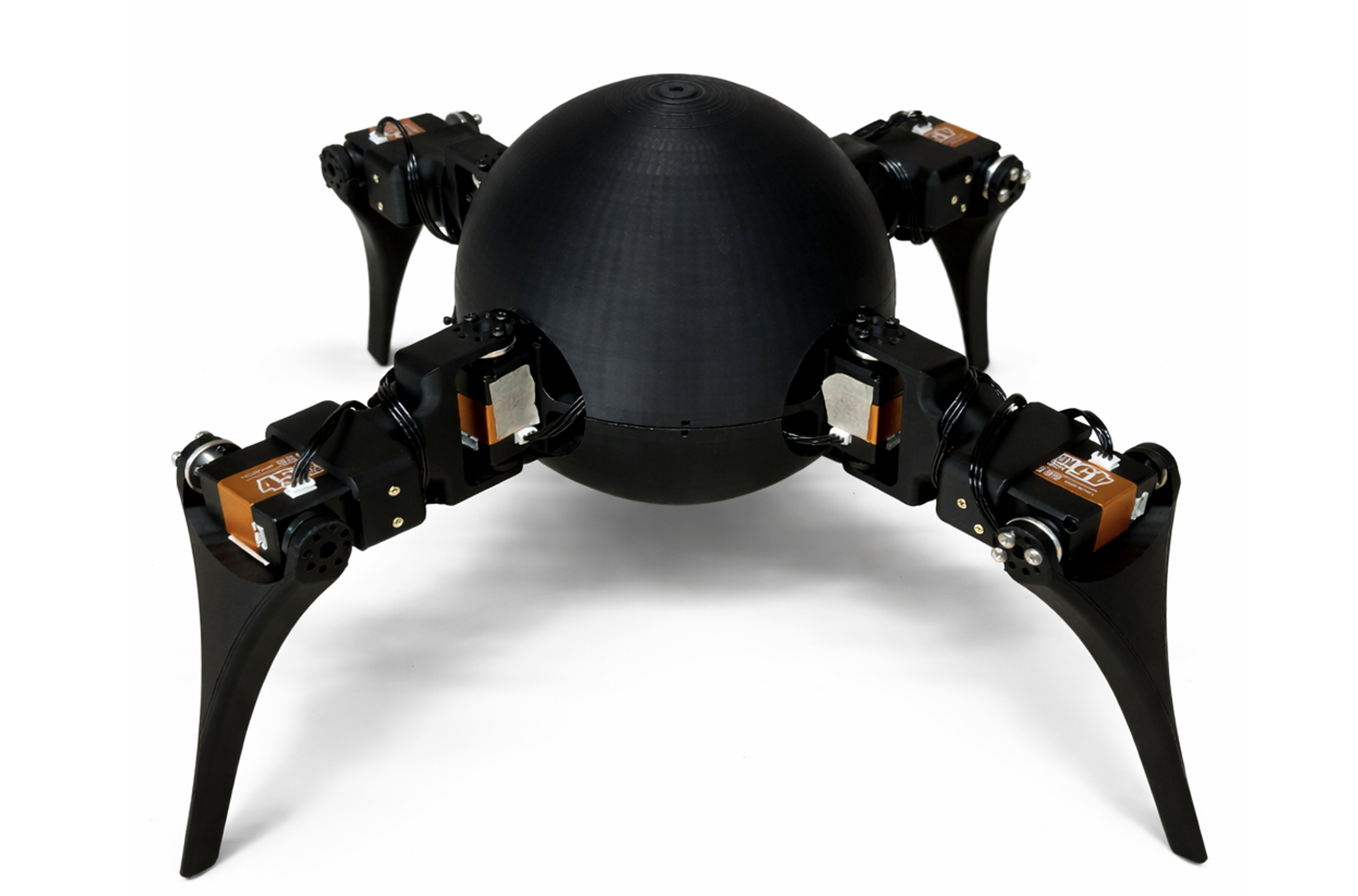}
\caption{Quadruped in simulation and hardware: (left) CAD model showing the disk-shaped torso and four 2-DOF legs, and (right) the physical robot used for trajectory replay.}
\label{fig:readyant-cad-real}
\end{figure}

Our quadruped was reverse engineered from the simulator: starting from the canonical Ant kinematic graph (floating torso, four legs, 8 actuated joints), we chose the simplest physical body that could replay Ant policies by matching key geometric ratios (hip-to-torso radius, proximal-to-distal link lengths) and centering the design on a disk-shaped torso with hip joints at fixed proportional offsets.

\subsection{Geometry and mapping from Ant}
\label{app:readyant-geometry}

We used the Gym/MuJoCo Ant as the reference morphology. In that model, the hip joints are placed at a radius that is approximately $1.13\times$ the torso radius, and each leg's distal link is roughly twice the length of the proximal link. Our quadruped preserved these two ratios while scaling the overall body down to a $160$\,mm diameter torso disk. Hip axes were placed at a $90$\,mm radius from the torso center, and the printed leg links were approximately $80$\,mm (hip--knee) and $160$\,mm (knee--foot). Table~\ref{tab:readyant-geometry} summarizes the resulting mapping.

\begin{table}[t]
\centering
\small
\caption{Key dimensions of the MuJoCo/Gym Ant and our Ant-like quadruped.}
\label{tab:readyant-geometry}
\begin{tabular}{lcc}
\hline
\textbf{Quantity} & \textbf{Ant (sim)} & \textbf{Our quadruped} \\
\hline
Torso radius & $0.25$\,m & $0.08$\,m \\
Torso diameter & $0.50$\,m & $0.16$\,m \\
Hip radius & $0.283$\,m & $0.09$\,m \\
Hip/torso radius & $\approx 1.13$ & $\approx 1.13$ \\
Upper leg length & $\approx 0.28$\,m & $0.08$\,m \\
Lower leg length & $\approx 0.57$\,m & $0.16$\,m \\
Upper:lower ratio & $\approx 1{:}2$ & $\approx 1{:}2$ \\
Leg reach (hip--foot) & $\approx 0.85$\,m & $\approx 0.24$\,m \\
\hline
\end{tabular}
\end{table}

To simplify integration, we reserved a central ``electronics safe'' region inside the torso for batteries and compute, and mounted batteries low to reduce interference and lower the center of mass.

\subsection{Hardware and bill of materials}
\label{app:readyant-bom}

Our quadruped was assembled from commodity parts:
\begin{itemize}
    \item \textbf{Actuators:} Hiwonder HTD-45H bus servos (eight used), with bus cables and onboard position feedback.
    \item \textbf{Servo interface:} A LewanSoul-style BusLinker controller board to connect the servo bus to the Raspberry Pi (with one spare available).
    \item \textbf{Compute:} Raspberry Pi 5 running the playback/control loop.
    \item \textbf{Power:} 3S LiPo batteries for the servos and compute, plus a buck converter for the Pi supply and basic power distribution hardware.
    \item \textbf{Structure:} 3D-printed torso disk, leg links, brackets, and feet (desktop FDM, PLA).
    \item \textbf{Fasteners:} M3 screws/nuts and heat-set inserts for servo and linkage mounting.
\end{itemize}

\subsection{Angle ``tape'' replay for retroactive sim-to-real}
\label{app:readyant-replay}

Trajectory replay used a recorded simulation rollout as a joint-angle ``tape.'' We took the first frame as a reference pose $q_0$, and at each timestep commanded the physical servos to track the relative motion $\Delta q(t)=q(t)-q_0$. Each joint's $\Delta q(t)$ (radians) was converted to servo encoder counts via a fixed counts-per-radian scale with a per-servo sign convention, then added to a calibrated per-servo neutral count that encoded the robot's physical reference posture. Playback speed was set by the streaming rate and move-time, and safety limits enforced per-step change bounds and absolute position windows around the neutral pose; optionally, the controller could wait for encoder feedback before advancing. This produced time-scaled position-setpoint replay of the simulated joint trajectories without requiring torque control.

%TC:endignore.
\end{document}